\documentclass[10pt]{article} 
\usepackage[preprint]{tmlr}

\usepackage{amsmath,amsfonts,bm}

\def\eqref#1{equation~\ref{#1}}

\def\1{\bm{1}}

\DeclareMathAlphabet{\mathsfit}{\encodingdefault}{\sfdefault}{m}{sl}
\SetMathAlphabet{\mathsfit}{bold}{\encodingdefault}{\sfdefault}{bx}{n}

\usepackage[utf8]{inputenc} 
\usepackage[T1]{fontenc}    
\usepackage{hyperref}       
\usepackage{url}            
\usepackage{booktabs}       
\usepackage{amsfonts}       
\usepackage{nicefrac}       
\usepackage{microtype}      
\usepackage{xcolor}         
\usepackage{amsfonts, amsmath, amsthm, bm, amssymb}
\usepackage{multirow}
\usepackage{mathtools}
\usepackage{enumerate}
\usepackage{longtable}
\usepackage[table]{xcolor}
\usepackage{makecell}
\usepackage{wrapfig}
\usepackage{float}
\usepackage{subcaption}
\usepackage[most]{tcolorbox}
\usepackage{enumitem}
\usepackage{array}
\usepackage{etoolbox}
\usepackage{lipsum}

\definecolor{lightyellow}{RGB}{255,255,204}
\definecolor{lightgreen}{RGB}{204,255,204}
\definecolor{lightblue}{RGB}{204,229,255}

\newcolumntype{C}[1]{>{\centering\arraybackslash}m{#1}}
\newcolumntype{L}[1]{>{\raggedright\arraybackslash}m{#1}}
\newcolumntype{R}[1]{>{\raggedleft\arraybackslash}m{#1}}

\definecolor{tabblue}{RGB}{217, 232, 252}
\definecolor{tabbluehead}{RGB}{184, 210, 247}
\definecolor{tabgreenhead}{RGB}{216, 235, 212}
\definecolor{tabyellowhead}{RGB}{244, 239, 196}

\definecolor{spbblue}{RGB}{44, 92, 152}
\definecolor{spbbluebg}{RGB}{245, 249, 255}
\definecolor{spbbluetitle}{RGB}{204, 221, 240}

\newtcolorbox{keyfindingsbox}[1][]{
    enhanced,
    breakable,
    colback=spbbluebg,
    colframe=spbblue,
    boxrule=0.6pt,
    arc=2mm,
    left=1.8mm,
    right=1.8mm,
    top=5mm,
    bottom=1.5mm,
    before skip=8pt,
    after skip=8pt,
    title=#1,
    coltitle=black,
    fonttitle=\bfseries,
    colbacktitle=spbbluetitle,
    attach boxed title to top left={
        xshift=2mm,
        yshift*=-\tcboxedtitleheight/2-0.5mm
    },
    boxed title style={
        boxrule=0.4pt,
        colframe=spbblue,
        arc=1.5mm,
        left=1.2mm,
        right=1.2mm,
        top=0.8mm,
        bottom=0.8mm
    }
}

\renewcommand{\thefootnote}{\fnsymbol{footnote}}

\title{SpikingBrain2.0: Brain-Inspired Foundation Models for Efficient Long-Context and Cross-Platform Inference}

\author{%
\name 
Yuqi Pan$^{1}$, 
Jinghao Zhuang$^{1}$, 
Yupeng Feng$^{1}$, 
Fangzhi Zhong$^{1}$, 
Siyu Ding$^{1}$, 
{Xuerui Qiu}$^{1}$,\\ 
{Shaowei Gu}$^{1}$, 
{Bohan Sun}$^{1}$, 
{Zhiyong Qin}$^{1}$, 
{Yibo Zhong}$^{1\ddagger}$, 
{Lingtao Ouyang}$^{1\ddagger}$, 
{Kun Yang}$^{1\ddagger}$,\\
{Zehao Liu}$^{1,4}$, 
{Yuhong Chou}$^{1,4}$, 
{Shurong Wang}$^{1}$, 
{Anjie Hu}$^{1}$, 
{Han Xu}$^{1}$, 
{Bo Xu}$^{1,2\dagger}$, 
{Guoqi Li}$^{1,2,3\dagger}$\\
~\\
\normalfont
$^{1}$Institute of Automation, Chinese Academy of Sciences\quad
$^{2}$Beijing Key Laboratory of Brain-Inspired General Intelligence Large Model\quad
$^{3}$Key Laboratory of Brain Cognition and Brain-inspired Intelligence Technology\quad
$^{4}$The Hong Kong Polytechnic University
}

\def\month{MM}  
\def\year{YYYY} 
\def\openreview{\url{https://openreview.net/forum?id=XXXX}} 

\begin{document}

\maketitle

\begingroup
\renewcommand{\thefootnote}{\fnsymbol{footnote}}
\footnotetext[2]{Corresponding authors: xubo@ia.ac.cn and guoqi.li@ia.ac.cn. Code: \url{https://github.com/BICLab/SpikingBrain2.0}.}
\footnotetext[3]{Work done during internship.}
\endgroup




\renewcommand{\thefootnote}{\arabic{footnote}} 
\setcounter{footnote}{0} 

\begin{abstract}
Scaling context length is reshaping the development paradigm of large models. However, conventional full-attention Transformers encounter prohibitive computational costs and inference bottlenecks as sequence lengths grow. A critical challenge remains: how to architect foundation models that sustain high performance and long-context efficiency while minimizing incremental training overhead. Building on the integration of brain-inspired mechanisms and efficient large-model design, we introduce \textbf{SpikingBrain2.0 (SpB2.0)}, a 5B model that substantially advances the architecture and training pipeline of its predecessor.
Our contributions are two-fold: 
\textbf{\romannumeral1) Architectural Innovation:} We propose Dual-Space Sparse Attention (DSSA), an inter-layer hybrid of Sparse Softmax Attention (MoBA) with Sparse Linear Attention (SSE). This brain-analogous sparse memory paradigm achieves a superior performance-efficiency trade-off for long-context modeling. 
In addition, SpB2.0 supports two quantization paths: INT8-Spiking coding enables sparse event-driven computation as an alternative to dense matrix multiplication, while FP8 coding targets practical acceleration on modern GPU platforms. 
\textbf{\romannumeral2) Enhanced Training Strategy:} We develop an optimized Transformer-to-Hybrid (T2H) conversion pipeline. By leveraging a curated set of open-source data, we implement dual conversion paths for LLMs and VLMs, facilitating seamless architectural migration.
Empirically, SpB2.0-5B and SpB2.0-VL-5B recover most of the capability of the base Transformer (Qwen3-4B) with a total training cost of fewer than 7k A100 GPU hours. SpB2.0 achieves a 10.13$\times$ TTFT speedup at 4M context length under sequence parallelism, and under vLLM it supports serving beyond 10M tokens on 8 A100 GPUs, where the full-attention baseline exceeds memory limits. SpB2.0 also shows strong cross-platform compatibility, enabling FP8 inference on conventional GPUs (2.52$\times$ speedup at 250k context) and asynchronous event-driven execution on neuromorphic chips (64.31\% spike-sequence sparsity, with 70.6\% and 46.5\% area and power reduction at 500MHz, respectively).
Overall, SpikingBrain2.0 offers a practical pathway for developing lightweight, multimodal, spiking hybrid models, validating the immense potential of synergizing brain-inspired mechanisms with efficient foundation architectures for resource-constrained and edge-computing environments.
\end{abstract}

\section{Introduction}
\label{sec:introduction}
Recent breakthroughs in large-scale foundation models have been driven not only by the scaling of data and parameters but also by the rapid expansion of context windows~\citep{anthropic2026claude1m,team2024gemini,liu2025deepseek,team2025kimi,blakeman2025nemotron,team2026minicpm}. As autonomous agents, codebase understanding, long-document analysis, and complex multimodal workflows become increasingly prevalent, efficient processing of long sequences has become a core requirement rather than an optional capability~\citep{liu2025comprehensive}. However, as sequence lengths scale, conventional full-attention Transformers face intrinsic limitations characterized by quadratic computational complexity and prohibitive memory overhead. This creates a significant bottleneck for deployment, particularly in resource-constrained environments or real-time applications. Consequently, a pivotal research frontier has emerged: how to architect foundation models that achieve both high-fidelity long-context modeling and superior inference efficiency, while maintaining a manageable training budget.

To address this challenge, we extend our exploration of integrating brain-inspired mechanisms with efficient large-model architectures and introduce \textbf{SpikingBrain2.0 (SpB2.0, Figure~\ref{fig:model_arch})}, a 5B hybrid spiking model series comprising SpB2.0-5B and SpB2.0-VL-5B. Compared to its predecessor~\citep{pan2025spikingbrain}, SpB2.0 offers substantial upgrades in both architectural design and training efficiency.

From the architectural perspective, the dominant computational bottleneck of standard Transformers shifts from feed-forward matrix multiplications to attention operations as sequence length increases. SpB2.0 mitigates this through two complementary designs:
\romannumeral1) \textbf{Dual-Space Sparse Attention (DSSA).} We introduce an inter-layer hybrid attention architecture, termed DSSA, that combines Sparse Softmax Attention (MoBA~\citep{lu2025moba}) and Sparse Linear Attention (SSE~\citep{pan2025scaling}) at a 1:3 ratio, aiming to better balance long-context capability and computational efficiency. Specifically, SSE operates sparsely over compressed state representations, while MoBA performs sparse computation over the exact, uncompressed KV cache. This brain-analogous sparsity of both mechanisms also helps mitigate attentional distraction and noise.
\romannumeral2) \textbf{Dual-Path Activation Coding}. To accelerate matrix multiplication (MatMul) across different hardware platforms and computing paradigms, we explore multiple quantization strategies. For event-driven neuromorphic hardware~\citep{roy2019towards,schuman2022opportunities,frenkel2023bottom}, we apply integer quantization together with spike-sequence expansion~\citep{xu2026spikedrivenlargelanguagemodel,xu2026spikemllmspikebasedmultimodallarge}, replacing dense MatMuls with sparse integer accumulation. For synchronous GPU execution, we adopt FP8 quantization to accelerate MatMuls by leveraging dedicated Tensor Cores on NVIDIA Hopper GPUs.

From the training perspective, continued training offers an efficient pathway for architectural migration under a limited training budget~\citep{kasai2021finetuning}, enabling substantial efficiency gains without extensive retraining. Accordingly, SpB2.0 adopts an efficient Transformer-to-Hybrid (T2H) conversion pipeline, with two dedicated paths for LLMs and VLMs, respectively: \romannumeral1) \textbf{LLM conversion.} We employ a multi-stage pipeline consisting of short-context distillation, three-stage long-context extension to 512k, general supervised fine-tuning (SFT), and reasoning-oriented SFT, supplemented by a preliminary exploration of on-policy distillation. Starting from Qwen3-4B-Instruct~\citep{yang2025qwen3} and using a fully open-source data mixture, the total training cost is approximately 4,656 A100 GPU hours (14B CPT tokens versus 150B in SpB1.0). \romannumeral2) \textbf{VLM conversion.} We adopt a two-stage pipeline consisting of knowledge distillation and instruction SFT. Starting from Qwen3-4B-VL-Instruction~\citep{bai2025qwen3} and using open-source datasets, the total training cost is approximately 1,977 A100 GPU hours.

Based on this architecture and conversion pipeline, we develop the SpikingBrain2.0 model series and demonstrate several key advantages:
\begin{itemize}
    \item \textbf{Long-Context Efficiency}:  \romannumeral1) Under HuggingFace sequence-parallel inference, SpB2.0 achieves up to 10.13$\times$ Time-to-First-Token (TTFT) speedup over the base Transformer at 4M context length.  \romannumeral2) Under vLLM tensor-parallel inference, it delivers 4.5$\times$, 1.12$\times$, and 4.3$\times$ speedups in TTFT, Time-Per-Output-Token (TPOT), and end-to-end latency, respectively, at 512k length. \romannumeral3) Owing to its reduced memory footprint, SpB2.0 also supports higher request concurrency and can serve sequences beyond 10M tokens on 8 A100 GPUs with chunked prefill.
    \item \textbf{Competitive Performance}: \romannumeral1) Despite relying solely on open-source data for lightweight conversion, SpB2.0 achieves performance comparable to Qwen3-4B on a broad set of general benchmarks (average $>$95\%) while consistently outperforming Qwen2.5-3B; its post-SFT performance also approaches that of Qwen3 and surpasses the larger SpB1.0-7B. \romannumeral2) In the multimodal setting, SpB2.0-VL attains performance comparable to strong full-attention baselines such as Qwen2.5-VL-3B, effectively recovering the overall capability of Qwen3-VL-4B.
    \item \textbf{Minimal Training Overhead}: The complete conversion of both the LLM and VLM models requires fewer than 7k A100 GPU hours in total and can be completed within 9 days on 32 A100 GPUs, while delivering competitive performance together with substantial efficiency gains. We further summarize several practical takeaways from this conversion process.
    \item \textbf{Cross-Hardware Compatibility}: By combining complementary quantization strategies, SpB2.0 can be deployed across diverse hardware platforms, including both GPUs and neuromorphic chips. \romannumeral1) The FP8 path achieves 2.52$\times$ TTFT speedup at 250k length with only 0.24\% accuracy loss. \romannumeral2) The INT8-Spiking path attains 64.31\% spike-sequence sparsity with only a 0.69\% performance drop, while simulation further shows significant hardware-efficiency gains over the INT8 baseline: area is reduced by 70.6\%, and power is reduced by 48.1\% and 46.5\% at 250MHz and 500MHz, respectively.
\end{itemize}

Overall, SpikingBrain2.0 provides a practical and lightweight path to develop multimodal, efficient spiking hybrid foundation models. More broadly, it further validates the potential of deeply integrating brain-inspired mechanisms with efficient large model architectures. Such designs are particularly appealing for long-context workloads, edge deployment, and resource-constrained scenarios, where efficient large-model inference is increasingly critical.

\section{Preliminary}\label{sec:preliminary}
This section introduces the background necessary for understanding SpikingBrain2.0. We first review attention mechanisms relevant to this work, including MoBA and SSE (Section~\ref{subsec:preliminary_attn}). We then describe the Transformer-to-Hybrid conversion paradigm (Section~\ref{subsec:t2h}), followed by a brief overview of spiking neural networks (Section~\ref{subsec:snn}). Finally, we revisit SpikingBrain1.0, the predecessor of this work, which provides an initial attempt to integrate brain-inspired mechanisms into large models (Section~\ref{subsec:spb1}). A summary of the notations used throughout the paper is provided in Appendix~\ref{app:notations}.

\subsection{Different Attention Mechanisms}\label{subsec:preliminary_attn}
Modern autoregressive foundation models can be viewed as sequence modeling systems. Inputs from different modalities are first tokenized into a sequence of representations $[\mathbf{x}_1,\dots,\mathbf{x}_t,\dots,\mathbf{x}_n]$, over which the model progressively refines latent features to perform downstream tasks. Within this framework, causal attention serves as the core mechanism for information interaction along the sequence dimension.

Given an input representation $\mathbf{x}_t$, the model projects it into a query, key, and value triplet. At time step $t$, the query $\mathbf{q}_t$ interacts with historical keys and values $\mathbf{k}_s,\mathbf{v}_s$ for $s\le t$ to produce the output representation $\mathbf{o}_t$.

\paragraph{Full Attention (FA)} In standard Transformers~\citep{vaswani2017attention}, the most common attention mechanism is softmax-based full attention, which computes exact pairwise interactions:

\begin{align}
    \mathbf{O}&=\operatorname{softmax}(\mathbf{Q}\mathbf{K}^{\top}+\log \mathbf{M})\mathbf{V}; \quad \mathbf{o}_t =\frac{\sum_{s=1}^t\operatorname{exp}(\mathbf{q}_t\mathbf{k}^{\top}_s)\mathbf{v}_s}{\sum_{s=1}^t\operatorname{exp}(\mathbf{q}_t\mathbf{k}_s^{\top})}.
\end{align}

Although FA provides precise sequence modeling, it becomes increasingly inefficient for long contexts. During training, it incurs quadratic computational complexity $O(n^2)$ (left formulation). During inference, memory usage grows linearly with sequence length because all historical KV pairs must be retained in the KV cache (right formulation). Consequently, pure FA becomes increasingly impractical as context length scales.

To address this limitation, two major families of efficient attention mechanisms have been widely studied: linear attention and sparse attention. They reduce the burden of long-context serving through different design principles and offer distinct trade-offs between efficiency and performance.

\paragraph{Linear Attention (LA)} Linear attention removes the softmax normalization from the attention map~\citep{katharopoulos2020transformers,sun2023retentive,yang2024gated,dao2024transformers,yang2024parallelizing}. In its recurrent form, historical KV information is compressed into a fixed-size state matrix $\mathbf{S}_t\in\mathbb{R}^{c\times d}$ through an outer-product update:

\begin{align}
    \mathbf{O}&=(\mathbf{Q}\mathbf{K}^{\top}\odot \mathbf{M})\mathbf{V}; \quad \mathbf{o}_t=\mathbf{q}_t\mathbf{S}_t, \ \text{where} \ \mathbf{S}_t=\mathbf{S}_{t-1}+\mathbf{k}_t^{\top}\mathbf{v}_t.
\end{align}

This formulation reduces both computational and memory overhead. During training, LA achieves linear complexity $O(n)$ (left formulation). During inference, it requires constant memory because the KV cache is replaced by the recurrent state matrix (right formulation). While LA scales favorably with sequence length, its limited state capacity often leads to degraded performance on long-context tasks that require complex reasoning or precise information retrieval~\citep{arora2024zoology,arora2024simple,jelassi2024repeat}.

\paragraph{Sparse State Expansion (SSE)} To alleviate the recall limitations of vanilla LA, SSE~\citep{pan2025scaling} expands the linear-attention state into multiple partitions with shared attention parameters. At each step, a gating vector $\mathbf{e}_t$ is computed, and only the top-$k$ partitions are selected for sparse state updates:

\begin{align}
    \mathbf{e}_t &= \operatorname{softmax}(\mathbf{x}_t\mathbf{W}_e), \quad \quad\mathcal{T} = \{i \ | \ \mathbf{e}_t^i \in \operatorname{top\text{-}}k(\mathbf{e}_t)\},\\
\mathbf{S}_{t}^i &= \begin{cases}
    \mathbf{S}_{t-1}^{i}+\mathbf{e}_t^i\cdot\mathbf{k}_t^{\top}\mathbf{v}_t, & \text{for}\ i\in \mathcal{T}\\
    \mathbf{S}_{t-1}^{i}, & \text{for}\ i\notin \mathcal{T}\\
\end{cases}\\
\mathbf{o}_t &= \sum_{i\in\mathcal{T}}\mathbf{e}_t^i\cdot\mathbf{q}_t\mathbf{S}_t^i.
\end{align}

To stabilize training of the sparse component, SSE introduces an always-selected partition and adopts a LoRA-based parameterization strategy. This design increases the effective state capacity of LA while keeping both computational and parameter overhead nearly constant, thereby improving long-context recall.

\paragraph{Sparse Attention (SA)} Unlike FA, sparse attention computes attention over only a subset of historical tokens rather than the entire KV history~\citep{xiaoefficient,li2024snapkv,tang2024quest,yuan2025native}. Such methods rely on selection strategies to identify the most relevant KV pairs. In this work, we focus on input-dependent block-sparse attention, with \textbf{Mixture of Block Attention (MoBA)}~\citep{lu2025moba} as a representative example. Specifically, the query $\mathbf{q}_t$ first interacts with block-pooled keys to produce selection scores, after which the top-$k$ KV blocks are chosen for exact attention computation:

\begin{align}
    \mathbf{e}_t &= \operatorname{softmax}(\mathbf{q}_t\operatorname{block\_pool}(\mathbf{K}_t)), \quad\mathcal{T} = \{i \ | \ \mathbf{e}_t^i \in \operatorname{top\text{-}}k(\mathbf{e}_t)\},\\
\mathbf{O} &= \operatorname{softmax}(\mathbf{Q}\mathbf{K}^{\top}_{i\in \mathcal{T}}+\log \mathbf{M})\mathbf{V}_{i\in \mathcal{T}}.
\end{align}

MoBA-style sparse attention reduces computation proportionally, resulting in subquadratic training complexity $O(n^2/b+nkb)$, where $b$ denotes the block size and $k$ denotes the number of selected blocks. However, because the full KV cache must still be preserved during inference, memory usage continues to grow linearly with sequence length.

\paragraph{Sliding-Window Attention (SWA)} SWA~\citep{beltagy2020longformer,child2019generating,Jiang2023Mistral7} is a special case of input-independent sparse attention. Instead of dynamically selecting historical tokens, each query attends only to a local window of preceding tokens:

\begin{align}
    \mathbf{o}_t =\frac{\sum_{s=t-w+1}^t\operatorname{exp}(\mathbf{q}_t\mathbf{k}^{\top}_s)\mathbf{v}_s}{\sum_{s=t-w+1}^t\operatorname{exp}(\mathbf{q}_t\mathbf{k}_s^{\top})}.
\end{align}

SWA reduces both computation and memory cost. Training complexity becomes linear in sequence length, and inference memory remains constant since only KV pairs within the local window are retained. However, similar to LA, SWA often suffers from substantial degradation in long-context scenarios, and for the same effective state size, this degradation can be even more severe~\citep{arora2024simple}.

\paragraph{Hybrid Attention} Hybrid attention combines multiple attention mechanisms within a single network to better balance efficiency and modeling capability. Two common paradigms are: \romannumeral1) inter-layer sequential hybridization~\citep{lieber2024jamba,rensamba,de2024griffin}, in which different attention mechanisms are assigned to different layers; and \romannumeral2) intra-layer parallel hybridization~\citep{dong2024hymba,zuo2025falcon,li2025transmamba}, in which multiple attention mechanisms operate in parallel within the same layer. SpikingBrain2.0 adopts a hybrid design that combines LA and SA, following a similar intuition to prior works~\citep{chen2021scatterbrain,hou2025rwkv,zhang2025sla,team2026minicpm}.

\subsection{Transformer-to-Hybrid Conversion} \label{subsec:t2h}
Most open-source large models are built on the Full Attention (FA) Transformer architecture, largely because of its mature ecosystem and stable training behavior. In addition, the majority of pretraining is typically conducted at relatively short context lengths (e.g., 4k or 8k tokens), under which standard FA Transformers can deliver strong performance without severe efficiency constraints. However, as model development progresses to later stages—such as long-context extension, post-training compute scaling, and inference serving—the cost of scaling context length gradually becomes a major bottleneck for Transformer architectures. This motivates a central question: how can a pretrained Transformer be efficiently converted into a more efficient hybrid model through continued training, while preserving its original capabilities and incurring minimal additional cost? This paradigm is commonly referred to as Transformer-to-Hybrid (T2H) conversion.

To improve the efficiency of T2H conversion, prior studies have explored multi-stage pipelines to increase training token efficiency. Early methods mainly relied on continual training over general-domain data~\citep{kasai2021finetuning,mercat2024linearizing,zhang2024gated}. Later work introduced a range of distillation techniques, including attention weights transfer~\citep{zhanghedgehog}, layer-wise output alignment~\citep{zhang2024lolcats}, and end-to-end logits-based knowledge distillation~\citep{wang2024mamba,bickllamba}, to accelerate convergence and improve capability recovery.

Empirical evidence also shows that the architectural gap between the source Transformer and the target hybrid model is a key factor in conversion efficiency. In particular, architectures that retain SWA, SA, or FA components tend to converge much faster than pure LA models~\citep{zhang2024lolcats,lan2025liger,tao2025infinitevl}. To further reduce this mismatch, several studies have also investigated hybrid layer arrangements~\citep{gu2025jet,yang2025zebra,li2025distilling}.

Recently, T2H-style pipelines have been adopted in many industrial-scale foundation models to improve long-context efficiency without extensive re-pretraining~\citep{liu2025deepseek,team2025every,team2026minicpm}, highlighting the practical effectiveness of this paradigm. SpikingBrain2.0 follows the same paradigm and improves T2H conversion through a combination of distillation strategies, hybrid layer arrangements, training data selection, and auxiliary SWA/FA modules.

\subsection{Spiking Neural Networks}\label{subsec:snn}
Traditional artificial neural networks (ANNs) primarily rely on dense matrix multiplication as their basic computational primitive. At the neuron level, this corresponds to computing weighted sums of inputs to produce output activations. In contrast, biological neurons exhibit substantially richer electrophysiological dynamics, which are often modeled by nonlinear differential equations~\citep{10636118}.

To better capture such dynamics, spiking neural networks (SNNs) are built from spiking neuron models such as the Leaky Integrate-and-Fire (LIF) neuron~\citep{maass1997networks,yao2023spike,meta_spikeformer}. A spiking neuron maintains an internal state, often referred to as the membrane potential, which integrates information over both spatial and temporal dimensions. When the membrane potential exceeds a predefined threshold, the neuron emits a spike to downstream neurons; otherwise, the potential gradually decays. This formulation is more biologically plausible and naturally supports event-driven computation, which can provide substantial energy-efficiency benefits during deployment. In particular, neurons update their states only when spikes are received and remain idle otherwise. As a result, SNNs deployed on neuromorphic hardware can achieve significantly lower energy consumption~\citep{davies2018loihi,roy2019towards,pei2019towards,schuman2022opportunities,frenkel2023bottom,Speck}.

However, scaling SNNs to large foundation models introduces additional requirements, including high parallelism to match modern GPU-based training infrastructure and sufficient information precision to support stable optimization. These requirements make conventional spiking neuron models, with their nonlinear recurrent dynamics and discrete spike transmission, poorly suited to large-scale training~\citep{luo2025integer,yao2024scaling,huang2026parallel}.

To address this limitation, recent work has explored spiking coding schemes that quantize activations while simplifying neuron dynamics in large-model settings~\citep{xingspikellm,pan2025spikingbrain,huang2026parallel}. By removing complex mechanisms such as nonlinear hard resets and detailed ion-channel dynamics, these approaches make computation more compatible with efficient GPU training. At the same time, they preserve the spiking property of activations, allowing activations to be expanded into sparse event sequences during inference and thereby retaining the energy-efficiency benefits of neuromorphic hardware. Under this design, part of the computation in large models can be shifted from dense MatMul operations to sparse additions. By controlling the sparsity of the event representation, such systems can in principle achieve substantial gains in energy efficiency, making them especially attractive for edge and neuromorphic deployment.

\subsection{SpikingBrain1.0}\label{subsec:spb1}
SpikingBrain1.0 (SpB1.0~\citep{pan2025spikingbrain}) is our initial effort to integrate brain-inspired mechanisms into large-scale models through a three-level co-design framework spanning model architecture, training pipeline, and system engineering. At the architectural level, SpB1.0 combines hybrid attention, Mixture-of-Experts (MoE) modules, and adaptive spiking coding. At the training level, it adopts an efficient conversion-based pipeline compatible with existing LLMs. At the system level, it includes customized training frameworks, operator libraries, and parallelization strategies tailored to the MetaX hardware platform.

Based on this framework, SpB1.0 develops two language models: SpikingBrain1.0-7B, a linear LLM, and SpikingBrain1.0-76B, a hybrid-linear MoE LLM. The primary objective of SpB1.0 is to improve long-context efficiency while moving toward more biologically inspired information processing paradigms. In particular, SpikingBrain1.0-7B achieves over 100$\times$ TTFT speedup on 4M-token inputs, together with more than 40$\times$ theoretical energy reduction relative to FP16 MAC-based computation, demonstrating its potential for energy-efficient large-model deployment.

\section{Model Architecture}\label{sec:model_arch}
\begin{figure}[t]
    \centering
    \includegraphics[width=0.92\textwidth]{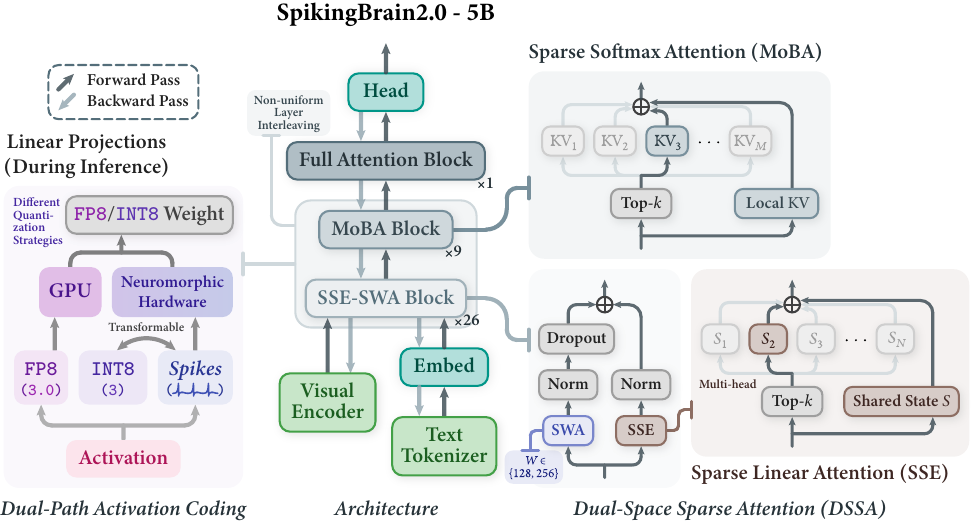}
    \caption{\textbf{Architecture of SpikingBrain2.0-5B (SpB2.0).} SpB2.0 adopts a 1:3 inter-layer hybrid design, termed DSSA, that combines MoBA and SSE, together with dual-path activation-coding strategies for linear projections. This design allows SpB2.0 to address the dominant computational bottlenecks of standard Transformers across different sequence-length regimes and hardware platforms.}
    \label{fig:model_arch}
    \vspace{-4mm}
\end{figure} 
Figure~\ref{fig:model_arch} illustrates the overall architecture of SpikingBrain2.0. At a high level, SpB2.0 adopts a Dual-Space Sparse Attention design, a non-uniform inter-layer hybrid of Sparse Softmax Attention (MoBA~\citep{lu2025moba}) and Sparse Linear Attention (SSE~\citep{pan2025scaling}) (Section~\ref{subsec:hybrid_model}), while retaining Full Attention (FA) only in the final layer. For the linear-attention layers, we further introduce an auxiliary SWA branch with a small window size, randomly dropped during training\footnote{This dropout is applied only during LLM training, whereas SpB2.0-VL retains the SWA branch throughout training.}. The feed-forward blocks follow the standard SwiGLU design. For linear projections, we incorporate multiple quantization strategies for activation coding to enable efficient MatMul computation across different hardware platforms and computing paradigms (Section~\ref{subsec:quant}).

\subsection{Design Considerations}
\label{subsec:trade_off}
\paragraph{Targeting Computational Bottlenecks Across Sequence Lengths}
In standard Transformer architectures, the dominant computational bottleneck shifts from feed-forward matrix multiplications (MatMuls) to attention operations as sequence length increases. Designing efficient solutions across different sequence-length regimes is therefore a central consideration in SpB2.0. To address this challenge, SpB2.0 adopts two complementary designs. First, it replaces full attention with an inter-layer hybrid attention architecture to improve efficiency under long contexts. Second, it introduces multiple quantization strategies, implemented as activation coding, to accelerate 16-bit dense MatMuls across different hardware platforms in short-context regimes. Together, these designs target the dominant bottlenecks in both short- and long-context settings, making SpB2.0 suitable for efficient deployment across diverse practical scenarios.

\paragraph{Toward a Better Long-Context Efficiency–Performance Trade-off}
Balancing modeling performance and computational efficiency under long contexts is another key consideration in our architectural design. In SpB1.0-7B, we adopted an inter-layer hybrid architecture composed of LA and SWA-4k. Its linear complexity provided excellent efficiency, but the limited state capacity of LA constrained long-context modeling. In SpB1.0-76B, we instead used a mixture of LA, SWA, and FA with carefully chosen layer ratios. Although this design improved long-context capability, the inclusion of multiple FA layers limited the overall efficiency gain.

To move toward a better operating point, SpB2.0 combines Sparse Attention (SA) with Linear Attention (LA), aiming to achieve both strong long-context modeling fidelity and high computational efficiency. Specifically, we retain only a single FA layer and pair Sparse State Expansion (SSE), a stronger long-context variant of vanilla LA, with SA modules to compensate for the performance loss that may arise from reducing FA layers. Prior studies~\citep{pan2025scaling} show that SSE-hybrids achieve higher recall after conversion than LA-hybrids, suggesting that stronger LA designs can reduce reliance on FA layers in hybrid architectures and thereby improve efficiency.

From the efficiency perspective, SSE, as an LA mechanism, reduces both computation and I/O, leading to lower latency and memory usage. MoBA, as an SA mechanism, reduces computation and I/O while retaining the same KV-cache storage footprint as FA. For the auxiliary SWA branch, we use a small window size comparable to the LA state size (i.e., 64-256 tokens), enabling highly efficient local attention computation. Overall, the SpB2.0 architecture provides a more favorable trade-off between long-context modeling performance and computational efficiency, enabling efficient long-context processing without substantially sacrificing precision.

\paragraph{Sparse Memory Modeling}
Importantly, both attention modules, SSE and MoBA, introduce intrinsic sparsity, which helps mitigate attention distraction caused by noisy long sequences~\citep{pan2025scaling}. Conceptually, SSE performs sparse computation over compressed state representations, whereas MoBA performs sparse computation over the exact, uncompressed KV cache. The two mechanisms are therefore naturally complementary and share a common sparse-memory intuition.

This sparse-memory design also aligns with brain-inspired memory mechanisms, which often rely on selective gating and competitive inhibition to regulate signal flow and suppress noise, as in interactions between the hippocampus and neocortex. In this sense, the hybrid sparse-memory design of SpB2.0 is not only computationally efficient, but also conceptually consistent with biologically analogous principles of memory organization.

\subsection{Hybrid Model Architectures with DSSA}
\label{subsec:hybrid_model}

\subsubsection{SpikingBrain2.0-5B Language Model}
\label{subsubsec:spb2_llm}
SpB2.0-5B language model is obtained by converting Qwen3-4B-Instruct-2507~\citep{yang2025qwen3} into a DSSA-based model via our Transformer-to-Hybrid pipeline, featuring an inter-layer hybrid of Sparse Softmax Attention and Sparse Linear Attention.

\begin{figure}[t]
    \centering
    \begin{minipage}[t]{0.5\textwidth}
        \centering
        \includegraphics[width=\textwidth]{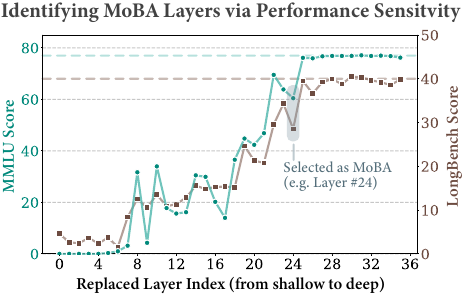}
    \end{minipage}
    \hfill
    \begin{minipage}[t]{0.47\textwidth}
        \centering
        \includegraphics[width=\textwidth]{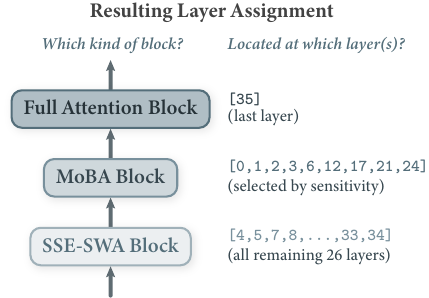}
    \end{minipage}
    \caption{\textbf{Layerwise performance sensitivity and resulting layer assignment for SpB2.0-5B.} \textbf{Left:} each point denotes the performance of a candidate model obtained by replacing a single FA layer with SSE. Dashed lines indicate the Qwen3 baseline performance on MMLU and LongBench. Layers whose replacement causes sharp degradation are selected as MoBA layers. \textbf{Right:} the resulting hybrid layer assignment, where the final layer is retained as FA, the sensitivity-selected layers are assigned to MoBA, and the remaining layers are converted to SSE-SWA.}
    \label{fig:layer_selection_curve}
    \vspace{-2mm}
\end{figure}

The model contains 36 layers (indexed from 0 to 35), of which 9 are replaced with MoBA attention, specifically layers [0,1,2,3,6,12,17,21,24]. These layers are selected using a greedy, training-free layer selection strategy. Concretely, we first replace the FA module in each layer with SSE individually, yielding 36 candidate models. We then evaluate all candidates in parallel on LongBench (32k) and MMLU. As shown in Figure~\ref{fig:layer_selection_curve}, performance degrades progressively as the replaced layer moves toward the bottom. This trend likely arises because earlier layers dominate residual information flow and their outputs propagate through all subsequent layers. Based on this observation, we analyze the performance curve from deeper to shallower layers and identify layers that exhibit sharp degradation. These layers are selected as MoBA layers, as replacing them with LA would substantially degrade context modeling. Empirically, this efficient training-free selection strategy yields better capability recovery than uniform layer interleaving, as shown in Table~\ref{tab:layer_selection_comparison}.

\begin{table}[t]
    \centering

    \begin{minipage}[t][5.6cm][t]{0.48\textwidth}
        \centering
        \captionof{table}{\textbf{Layer selection comparison.} Selected layers outperform uniform interleaving after SFT with 100k samples on MMLU, LB-32k (LongBench 32k), and IFEval.}
        \label{tab:layer_selection_comparison}
        \footnotesize
        \setlength{\tabcolsep}{4pt}
        \resizebox{\linewidth}{!}{
        \begin{tabular}{l|cccc}
            \toprule
             & Train loss & MMLU & LB-32k & IFEval \\
            \midrule
            Uniform  & 0.875 & 55.23 & 13.65 & 40.53 \\
            \textbf{Selected} & \textbf{0.846} & \textbf{60.32} & \textbf{14.06} & \textbf{47.72} \\
            \bottomrule
        \end{tabular}
        }
    \end{minipage}
    \hfill
    \begin{minipage}[t][5.6cm][t]{0.48\textwidth}
        \centering
        \captionof{table}{\textbf{SSE configuration comparison.} Under N4k2 settings after SFT with 100k samples, SSE-GDN shows better convergence than SSE-GLA, while the ReLU feature map brings no consistent gains.}
        \label{tab:sse_configuration_comparison}
        \footnotesize
        \setlength{\tabcolsep}{4pt}
        \begin{tabular}{l|cc}
            \toprule
             & Train loss & MMLU \\
            \midrule
            SSE-GLA      & 0.955 & 44.97 \\
            SSE-GLA-ReLU & 0.990 & 45.69 \\
            \textbf{SSE-GDN}      & \textbf{0.871} & \textbf{49.67} \\
            SSE-GDN-ReLU & 0.880 & 48.34 \\
            \bottomrule
        \end{tabular}
    \end{minipage}
    \vspace{-1.8cm}
\end{table}

In addition to these MoBA layers, we retain the final layer (layer 35) as FA. This design is motivated by observations from MoBA~\citep{lu2025moba}, which suggest that sparse attention may lead to suboptimal SFT performance due to sparse gradient propagation. Retaining the final layer as FA helps stabilize training.

The remaining 26 layers are converted to SSE. During conversion, we replace the original GQA with MHA by repeating the KV projections. Empirically, we find that compressed linear attention suffers more severe recall degradation under GQA than softmax attention, likely because LA performs retrieval through a single matrix multiplication and therefore has weaker query expressiveness. Specifically, we adopt SSE-GDN, which consistently yields better post-conversion performance than SSE-GLA (Table~\ref{tab:sse_configuration_comparison}). For the QK feature map, we use SiLU, as our experiments show that the widely used non-negative ReLU feature map does not improve capability recovery (Table~\ref{tab:sse_configuration_comparison}). We also exclude short convolutions, as they further increase architectural deviation from the base Transformer.

For SSE layers, we introduce an auxiliary SWA branch with window size 128 and randomly drop it during training with a dropout rate of 0.5. This design is motivated by the complementary properties of LA and SWA. Under a comparable state size, pure LA offers stronger out-of-window retrieval but converges more slowly, whereas small-window SWA accelerates convergence at the cost of recall. Following prior work~\citep{benfeghoul2025untangling}, we therefore apply dropout to the SWA branch to reduce interference with the main LA pathway and avoid computational shortcuts that may weaken LA learning. The outputs of the two branches, after output projection, are first normalized with RMSNorm, which we term \emph{merge norm}, and then summed. This design aligns activation scales and improves stability during the early stage of conversion training. As shown in Table~\ref{tab:sse_swa_ablation}, the combination of these techniques, including a small SWA window, branch dropout, and merge norm, effectively prevents degradation of the LA branch. In particular, when evaluated with SSE-only inference, the model retains most of the capability of the full model, whereas SWA-only inference leads to a substantial performance drop. 

Finally, MoBA, FA, and SWA retain RoPE and QK-Norm, while SSE-GDN does not use positional encoding and instead applies L2 normalization to the query and key vectors.

\subsubsection{SpikingBrain2.0-5B Vision-Language Model}
\label{subsubsec:spb2_vlm}
SpikingBrain2.0-VL-5B (SpB2.0-VL) is obtained by converting Qwen3-4B-VL-Instruction~\citep{bai2025qwen3}. Its language backbone follows the same architecture as SpB2.0-5B, including the hybrid attention design, layer arrangement, and detailed SSE configuration. The visual encoder is kept identical to that of Qwen3-VL, and visual features are injected into the language backbone through the DeepStack feature injection mechanism~\citep{meng2024deepstack}.

Compared with the LLM configuration, several adjustments are introduced to better support multimodal tasks. Specifically, for the language backbone of the VLM, the auxiliary SWA branch uses a larger window size of 256, random dropout on the SWA branch is disabled during training, and a larger top-$k$ value is used for MoBA, as discussed in Section~\ref{subsec:vlm_conversion}.

These modifications are motivated by the observation that vision-language tasks typically require more precise and more global token interactions than pure language modeling. As a result, VLMs rely more heavily on dense token-level interactions, which are more naturally captured by softmax-based attention mechanisms.

\subsection{Quantization and Activation Coding}
\label{subsec:quant}
\begin{figure}[t]
    \centering
    \includegraphics[width=0.95\textwidth]{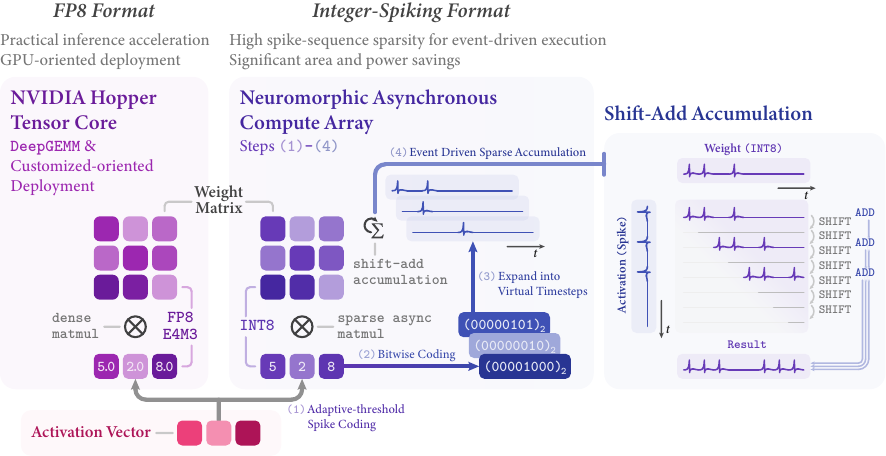}
    \caption{\textbf{Dual quantization paths in SpikingBrain2.0.} The FP8 path targets practical inference acceleration by executing FP8 MatMul on NVIDIA Hopper Tensor Cores. The INT8-Spiking path converts activations into sparse spike sequences, enabling sparse event-driven accumulation on asynchronous neuromorphic hardware. Together, the two paths support efficient deployment across both mainstream GPU platforms and neuromorphic systems.}
    \label{fig:quant}
\end{figure}

To support diverse hardware platforms and computing paradigms, SpikingBrain2.0 adopts two complementary quantization paths, whose overall design is illustrated in Figure~\ref{fig:quant}. The INT8-Spiking path preserves and validates the brain-inspired properties of our framework, including adaptive-threshold spiking and event-driven computation, with minimal accuracy loss. In parallel, the FP8 path targets contemporary GPU infrastructure and provides strong practical efficiency. Together, these two quantization strategies enable SpB2.0 to bridge biologically inspired spiking computation with efficient deployment on modern hardware.

\subsubsection{Integer–Spiking Format for Neuromorphic Hardware}
\label{subsubsec:int_spike_quant}
As the biologically inspired cornerstone of SpikingBrain2.0 (SpB2.0), the Integer-Spiking quantization path inherits and extends the spiking coding mechanism of SpB1.0, formalizing it into a quantization-native framework that bridges integer arithmetic with event-driven neuromorphic computation. This design is rooted in a key insight: integer activations and sparse spiking events are mathematically isomorphic, such that an integer value can be directly mapped to the cumulative count of spiking events over time.

Specifically, integer-valued activations, obtained through Adaptive-threshold Spiking Coding from SpB1.0~\citep{pan2025spikingbrain} (Figure~\ref{fig:quant}, step~(1)), can be expanded into multi-timestep sparse event sequences through dedicated encoding schemes such as bitwise coding~\citep{xu2026spikemllmspikebasedmultimodallarge} (Figure~\ref{fig:quant}, steps~(2)-(3)): each bit in the 8-bit integer represents the firing potential (1) or resting potential (0) at the corresponding virtual timestep. This encoding ensures that the expanded event sequence retains high sparsity, as most bits in quantized integers are zero for typical inputs, a property that is also consistent with biological neural systems, where most neurons remain inactive at a given moment. Such sparsity is critical for neuromorphic hardware, which leverages asynchronous event-driven processing to skip computations at non-spiking timesteps, thereby substantially reducing energy consumption compared to dense synchronous computation (Figure~\ref{fig:quant}, step~(4)). For the implementation of quantization, we adopt a fine-grained recipe:

\begin{itemize}[leftmargin=1.5em, itemsep=0.35em, topsep=0.35em]
    \item \textbf{Weights}: 128$\times$128 block-level INT8 quantization. For each block, we optimize the clipping coefficient via greedy search to minimize the local mean squared error (MSE) between full-precision and quantized weights. This block-wise design balances quantization accuracy and computational efficiency.
    
    \item \textbf{Activations}: 1$\times$128 group-level INT8 quantization. The computation of each group is equivalent to implementing the Adaptive-threshold Spiking Neuron mechanism, and with bitwise coding, each INT8 value can be converted into a signed 7-step spike sequence. This finer-grained grouping is more conducive to preserving quantization accuracy while retaining the core adaptive-threshold logic, avoiding over-excitation or quiescence without relying on explicit decay factors typical in neuromorphic modeling.
\end{itemize}

This INT8-Spiking pipeline strictly preserves the biological plausibility of SpB1.0, including adaptive spiking dynamics and event-driven computation, while achieving a mere 0.69\% accuracy drop relative to the full-precision baseline shown in Section~\ref{subsubsec:capability_perf_quant}. Unlike dense floating-point matrix multiplications in traditional LLMs, our framework can replace core computations with sparse integer additions with suitable neuromorphic hardware—each operation corresponds to a spiking event's contribution, inspired by the energy-efficient signal propagation in biological brains.

Designed explicitly for emerging neuromorphic hardware, this path enables seamless deployment on asynchronous event-driven architectures, reducing hardware overhead while maximizing energy efficiency. When combined with the GPU-optimized FP8 path, it allows SpB2.0 to span from biologically inspired neuromorphic systems to mainstream computing platforms, delivering a versatile solution for both fundamental brain-inspired research and practical LLM deployment.

\paragraph{Hardware Implementation} To support Integer–Spiking Format computation, we design a dedicated hardware architecture that maps dense INT8 matrix multiplication to sparse Integer–Spiking matrix computation. The architecture consists of independent on-chip data/weight buffers, a 128$\times$128 compute array, an accumulator, and a control module. The compute array consists of 128 parallel 128-input pipelined adder trees. Each adder tree has an independent weight input while sharing activation inputs. At runtime, the control module reads weights and activations, applies bitwise coding to generate spike sequences, and feeds them serially into the compute units. The compute units perform sparse accumulation. The accumulator uses shift-add across temporal steps and directly sums results across tiled blocks. The measured area and practical power results are reported in Section~\ref{subsubsec:efficiency_perf_quant}.

\subsubsection{FP8 Format for Industrial GPUs}
\label{subsubsec:fp_quant}
Since neuromorphic hardware is not yet widely available, most LLM development and deployment still rely on GPUs operating in a synchronous computing paradigm. To evaluate efficiency on modern industrial hardware, we develop a GPU-friendly FP8 variant of SpB2.0. Specifically, we implement the activation coding path in the FP8 E4M3 format and leverage Tensor Cores on NVIDIA Hopper GPUs using DeepGEMM and customized Triton kernels. These kernels support weight and activation quantization, fused SwiGLU with quantized output, and fused RMSNorm with quantized output. We further replace the standard self-attention operator with SageAttention2~\citep{zhang2025sageattention2} to improve efficiency. This FP8 implementation serves as a practical performance proxy for modern GPU platforms, while reducing the accuracy gap to only 0.24\%, as shown in Section~\ref{subsec:quant_perf}.

At the same time, several small projection components in SSE, such as QK-LoRA and selective gating, as well as MoBA quantization, exhibit relatively high implementation complexity but unfavorable gain-overhead ratios. These components are therefore retained in BF16 precision to preserve computational fidelity.

\section{Training Pipeline}
\label{sec:training_pipeline}
\begin{figure}[t]
    \centering
    \includegraphics[width=0.9\textwidth]{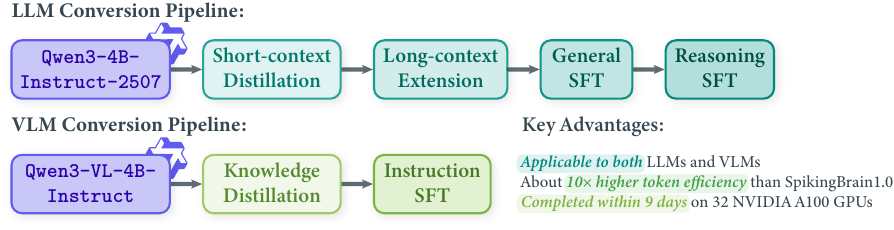}
    \caption{\textbf{Training pipelines for Transformer-to-Hybrid (T2H) conversion in SpikingBrain2.0.} SpB2.0 adopts dedicated conversion paths for LLMs and VLMs, enabling efficient architectural migration across both language-only and vision-language settings.}
    \label{fig:t2h_conversion}
\end{figure}

The objective of Transformer-to-Hybrid (T2H) conversion in SpikingBrain2.0 is to recover model capability at minimal additional training cost while remaining applicable to both language-only and vision-language models. This design enables substantial efficiency gains without extensive retraining. As illustrated in Figure~\ref{fig:t2h_conversion}, we develop two dedicated conversion pipelines tailored for LLMs (Section~\ref{subsec:llm_conversion}) and VLMs (Section~\ref{subsec:vlm_conversion}). Compared with SpB1.0, the proposed framework extends to a broader modality setting and improves token efficiency.

\subsection{LLM Conversion}
\label{subsec:llm_conversion}
We initialize SpB2.0-5B from Qwen3-4B-Instruct-2507. Most parameters are inherited from the base Transformer, while newly introduced parameters are randomly initialized. As summarized in Table~\ref{tab:training_config_llm}, we then perform a multi-stage conversion pipeline consisting of short-context distillation, long-context extension to 512k, general supervised fine-tuning (SFT), and reasoning-oriented SFT. Across all stages, we carefully control the training token budget to ensure effective capability recovery while maintaining high training efficiency. 
In addition, we conduct a preliminary study on on-policy distillation, providing initial evidence for the feasibility of applying RL-style optimization to SpB2.0.

The total training cost of SpB2.0-5B conversion is approximately 4,656 A100 GPU hours. As a result, the full development pipeline, from the base pretrained model to the long-context and instruction-tuned versions, can be completed within one week on 32 A100 GPUs.

\begin{table}[t]
    \centering
    \caption{\textbf{Training configurations for each stage of LLM T2H conversion.} G/R-SFT: general/reasoning SFT; LR: learning rate; Const/Cos: constant/cosine-decay learning-rate schedule.}
    \label{tab:training_config_llm}
    \small
    \resizebox{0.85\linewidth}{!}{
    \begin{tabular}{l|cccccc}
        \toprule
         & \textbf{Distillation} & \textbf{LongCT1} & \textbf{LongCT2} & \textbf{LongCT3} & \textbf{G-SFT} & \textbf{R-SFT} \\
        \midrule
        Peak LR & $3\times10^{-5}$ & $3\times10^{-6}$ & $3\times10^{-6}$ & $3\times10^{-6}$ & $5\times10^{-6}$ & $5\times10^{-6}$ \\
        End LR & $3\times10^{-6}$ & $3\times10^{-6}$ & $3\times10^{-6}$ & $3\times10^{-6}$ & $5\times10^{-7}$ & $5\times10^{-7}$ \\
        LR schedule & Cos & Const & Const & Const & Cos & Cos \\
        \midrule
        Optimizer & \multicolumn{6}{c}{AdamW ($\beta_1=0.9$, $\beta_2=0.95$)} \\
        Weight decay & \multicolumn{6}{c}{0.1} \\
        Grad clip & \multicolumn{6}{c}{1.0} \\
        \midrule
        Sequence length & 8k & 64k & 256k & 512k & 8k & 24k \\
        Batch size (samples) & 64 & 16 & 4 & 2 & 64 & 32 \\
        Training tokens (B) & 3 & 3 & 3 & 5 & \textbackslash & \textbackslash \\
        Training samples & 400k & 50k & 12k & 9.6k & 500k & 500k \\
        \midrule
        Auxloss weight & \multicolumn{6}{c}{0.001} \\
        SWA dropout rate & \multicolumn{6}{c}{0.5} \\
        \midrule
        MoBA chunk size & 512 & 1k & 4k & 4k & 4k & 4k \\
        MoBA topk & 4 & 8 & 12 & 12 & 12 & 12 \\
        \midrule
        GPU hours of A100 & 400 & 448 & 608 & 1536 & 384 & 1280 \\
        \bottomrule
    \end{tabular}
    }
\end{table}

\paragraph{Short-context distillation} 
Because the architecture changes substantially after conversion, resulting in an initial training loss above 10, we first perform distillation on high-quality generic data using the original Transformer as the teacher. This stage aims to rapidly recover the model's general language modeling capability while enabling a smooth adaptation to the target architecture. We combine layer-wise output alignment via MSE with end-to-end logits-based knowledge distillation, using the following objective:
\begin{align}
    \mathcal{L} &= 
\mathcal{L}_{\text{CE}} + c\cdot 
\mathcal{L}_{\text{aux}} + \alpha \cdot 
\mathcal{L}_{\text{KD}} + \beta \cdot |
\mathcal{L}_{\text{KD}} / 
\mathcal{L}_{\text{MSE}}|  \cdot 
\mathcal{L}_{\text{MSE}}.
\end{align}

The definitions of these loss terms are provided in Appendix~\ref{app:loss_def}. During distillation, we set $\alpha=\beta=0.1$, which empirically performs better than setting both coefficients to 1. For the KD loss, we select the top-128 logits and renormalize them to form the teacher distribution. For the MSE term, we average the alignment loss across all layers using attention outputs. The MoBA selection ratio is set to 25\%, corresponding to 2k selected tokens.

\begin{wrapfigure}{r}{0.52\textwidth}
    \centering
    \vspace{-4.5mm}
    \includegraphics[width=\linewidth]{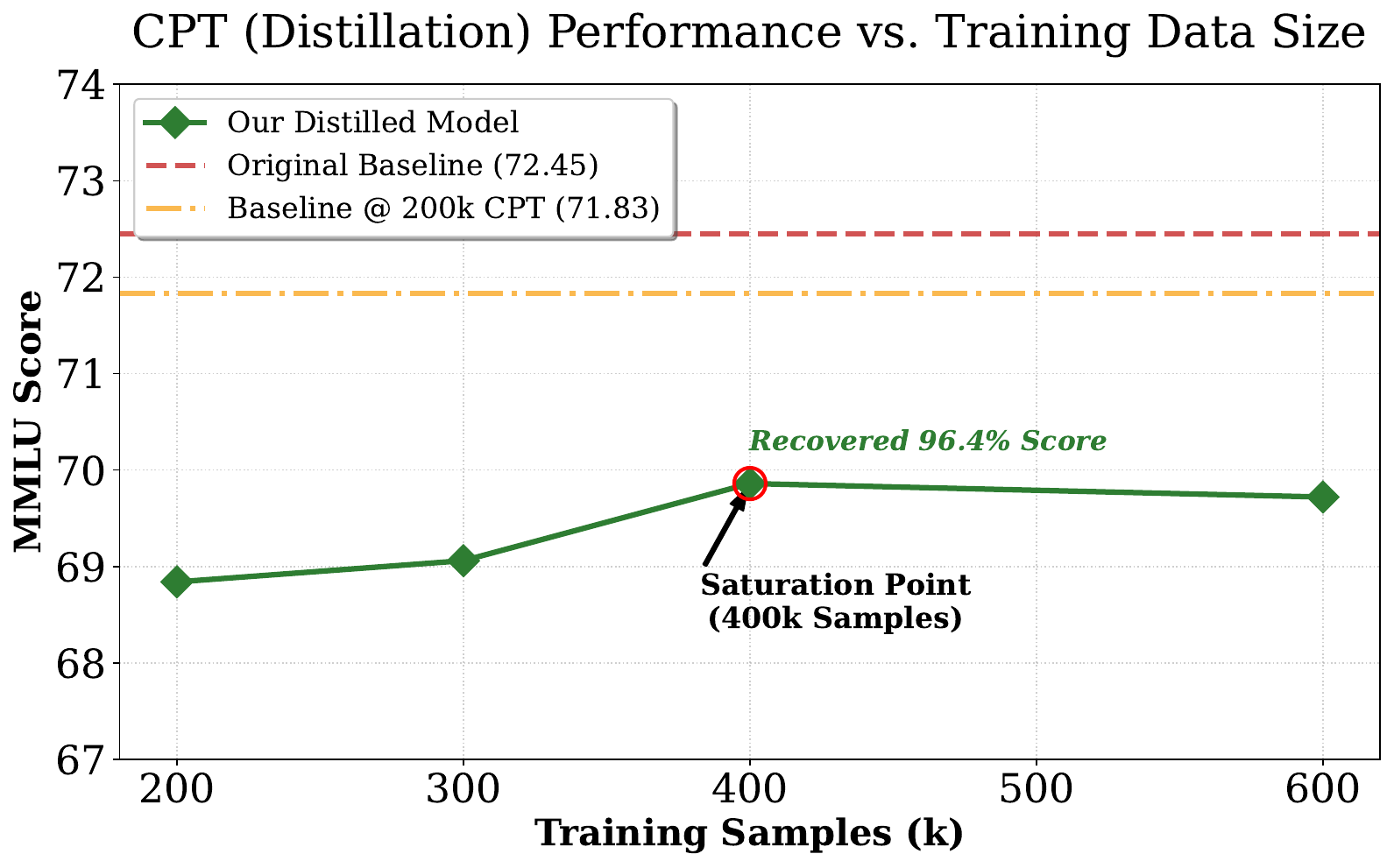}
    \vspace{-4mm}
    \caption{\textbf{MMLU score versus distillation training data size.} Dashed lines indicate the Qwen3 MMLU baselines, including the original score and the score after training on our 200k CT dataset.}
    \label{fig:cpt_data_saturation}
    \vspace{-3mm}
\end{wrapfigure}

At this stage, we train on a high-quality continual-training mixture composed of Nemotron-Pretraining-SFT-v1~\citep{basant2025nvidia} and a high-quality subset of Nemotron-CC-v2~\citep{basant2025nvidia} (sampled from Diverse-QA and High-Quality) at a ratio of 1:3. All short-context samples are packed into 8k sequences. In total, we use 400k samples (approximately 3B tokens) for distillation. We further evaluate dataset sizes from 200k to 600k samples and find that 400k is an effective turning point, beyond which performance gains begin to saturate, as shown in Figure~\ref{fig:cpt_data_saturation}. This result suggests that most knowledge transfer has already been completed. After this stage, the model recovers more than 95\% of the original MMLU score (69.86 vs.~72.45).

\paragraph{Long-context extension}
After short-context distillation, we conduct progressive long-context continual training to restore the long-context capability of the target hybrid model. Even when the base Transformer already supports long contexts, conversion followed only by short-context CPT is insufficient to recover this capability; high-quality long-context data are still required to reactivate it. We therefore adopt three stages of progressive length expansion: 64k, 256k, and 512k. During training, we jointly increase the MoBA chunk size and top-$k$ value, resulting in activation ratios of 12.5\%, 18.8\%, and 9.4\%, respectively. Concurrently, the sequence-parallel degree is increased from 4 to 8 and then to 16 using Ulysses~\citep{jacobs2023deepspeed}.

Unlike SpB1.0, which relies on unmasked concatenated short texts, we train on naturally long documents with variable lengths. Specifically, we use the ProLong~\citep{gao2025train} dataset and replace its short-text mixture (originally from FineWeb-Edu) with the same high-quality Nemotron-CC subset used in the previous stage. Across the three stages, we train on 11B tokens with a constant learning rate, including 5B tokens at the 512k stage. Compared with a 3B-token variant, this setting yields similar pretrained performance but noticeably better post-SFT performance on tasks such as LongBench and GSM8K. As shown in Table~\ref{tab:longct_stage_comparison}, long-context capability improves steadily across the three stages, while short-context benchmarks such as MMLU decrease only slightly.

\paragraph{General SFT}
After long-context training, we perform general SFT to strengthen instruction-following (IF) capability. This stage does not include explicit CoT data. We observe that, even when converting from an instruction-tuned base model, IF performance degrades substantially after continual training if no SFT is applied. This degradation is particularly visible on generative benchmarks such as LongBench.

To improve data diversity, we mix several SFT datasets covering chat, mathematics, and code tasks. Training is conducted at an 8k sequence length for one epoch over 500k samples. Increasing the number of epochs does not yield meaningful downstream gains. We also experiment with long-context checkpoint merging, which slightly improves pretrained performance but does not translate into consistent gains after SFT.

\paragraph{Reasoning-oriented SFT}
To equip the model with explicit CoT reasoning capability, we perform an additional reasoning-oriented SFT stage after general SFT. We use a mixture of datasets covering mathematics, code, science, dialogue, and safety tasks. Training is conducted with a 24k sequence length, as more than 99\% of the collected samples fall within this range, using 500k samples for one epoch. After this stage, the model is able to generate explicit reasoning traces and achieves improved performance on reasoning-intensive tasks such as mathematics and coding.

\paragraph{On-policy Distillation (OPD)} 
Following Reasoning SFT, we investigate on-policy distillation~\citep{agarwal2024policy,lu2025onpolicydistillation} to further enhance complex reasoning. We employ Qwen3-4B-Thinking-2507 as the teacher and the non-RL SpB2.0-5B as the student, performing distillation on the deduplicated DAPO-Math-17k dataset with a reverse KL objective. During this process, the student generates responses on-policy and aligns its logits with those of the teacher. OPD is particularly effective in this setting for two primary reasons: first, the T2H conversion ensures vocabulary consistency between the teacher and student; second, the Qwen3-Thinking model has undergone additional post-training stages, exhibiting superior reasoning ability (e.g., higher AIME performance) and thus serving as a high-quality distillation target. Compared to vanilla RL-based methods such as GRPO~\citep{shao2024deepseekmath} and DAPO~\citep{yu2025dapo} that rely on sparse rewards, we find that OPD yields more stable performance improvements.

Training is implemented using the verl framework on 16 A100 GPUs, with vLLM for rollout and FSDP as the training backend, together with the one-step-wise strategy for computational efficiency. We set the logits Top-$k$ to 64 and cap the maximum response length at 9k tokens. The model is trained for 290 steps (approximately 18k samples). The KL loss converges stably, and the resulting model shows clear improvements in advanced mathematical reasoning and coding tasks. These results provide preliminary evidence for the efficacy of RL-style optimization on SpB2.0, with larger-scale RL exploration left for future work.

\begin{keyfindingsbox}[Key Findings from LLM T2H Conversion]
During the development of the LLM T2H pipeline, we conducted extensive experiments and obtained several practical insights beyond simple hyperparameter tuning.

\begin{itemize}[leftmargin=1.5em,itemsep=0.35em,topsep=0.35em]
    \item \textbf{A CPT warm-up stage is essential, and pure SFT-based conversion cannot effectively recover model capability.} Unlike some prior studies~\citep{zhang2024lolcats,wang2024mamba}, we find that pure SFT yields only limited recovery of general capabilities, often accompanied by noticeable performance degradation and poor token efficiency. Empirically, 100k SFT samples achieve performance comparable to only 10k CPT samples. We hypothesize that the learning signal provided by SFT is too sparse for architectural conversion, making a CPT warm-up stage necessary.

    \item \textbf{Distillation is much more effective during CPT than during SFT.} Although distillation can reduce SFT loss, directly applying distillation in the SFT stage does not consistently improve downstream performance and may even degrade it. In contrast, distillation during CPT consistently improves benchmark performance.

    \item \textbf{Data quality is critical for CPT.} High-quality data subsets yield substantially better results than generic pretraining data. We evaluate multiple datasets, including DCLM, Nemotron-CC, and its high-quality subsets, and find that higher-quality data significantly accelerates capability recovery under a limited training budget.
    
    \item \textbf{General-domain capability can be largely recovered with a surprisingly small token budget.} In SpB1.0, we used roughly 100B tokens for CPT, but our experiments reveal considerable room for compression. In SpB2.0, using only 1.5B and 3B CPT tokens already recovers 94.3\% and 96.4\% of the original MMLU score, respectively. Achieving this level of recovery, however, depends on both high-quality data and accurate layer selection.

    \item \textbf{Long-context training is necessary after conversion.} Training only on short sequences restores short-context capability but fails to generalize to longer sequence lengths, even when the base model already supports long contexts. Proper long-context training stages are therefore required to activate the hybrid model's long-context capability.
\end{itemize}
\end{keyfindingsbox}

\subsection{VLM Conversion}
\label{subsec:vlm_conversion}
Unlike conventional VLM construction, which typically combines a pretrained LLM with a vision encoder~\citep{liu2023visual,lillava}, SpikingBrain2.0-VL (SpB2.0-VL) is obtained through T2H conversion from Qwen3-4B-VL-Instruction. As a result, SpB2.0-VL inherits a certain level of the visual understanding capability of the original model, allowing us to avoid large-scale vision-language alignment training from scratch. Instead, the main challenge is to mitigate the capability perturbation introduced by architectural conversion.

During conversion, we inherit most parameters from Qwen3-VL, including the vision encoder, while randomly initializing the newly introduced modules. To restore and stabilize model performance after conversion, we adopt a two-stage training pipeline consisting of knowledge distillation followed by instruction supervised fine-tuning (Instruction SFT), as summarized in Table~\ref{tab:training_config_vlm}.

The total training cost of SpB2.0-VL conversion is approximately 1,977 A100 GPU hours. As a result, the full development pipeline can be completed within one week on 16 A100 GPUs.

\begin{table}[ht]
    \centering
    \caption{{\textbf{Training configurations for each stage of VLM T2H conversion.} KD: knowledge distillation; VE: vision encoder; LR: learning rate; Cos: cosine-decay learning-rate schedule.}}
    \label{tab:training_config_vlm}
    \small
    \resizebox{0.5\linewidth}{!}{
    \begin{tabular}{l|cc}
        \toprule
         & \textbf{KD} & \textbf{instruct-SFT} \\
        \midrule
        Backbone LR & $5\times10^{-5}$ & $5\times10^{-5}$ \\
        VE LR & $1\times10^{-5}$ & $1\times10^{-5}$  \\
        LR schedule & Cos & Cos \\
        \midrule
        Optimizer & \multicolumn{2}{c}{AdamW ($\beta_1=0.9$, $\beta_2=0.99$)} \\
        Weight decay & \multicolumn{2}{c}{0.1} \\
        Grad clip & \multicolumn{2}{c}{1.0} \\
        \midrule
        Max sequence length & 8k & 8k \\
        Packing length & \textbackslash & 16k \\
        Batch size (samples) & 512 & \textbackslash \\
        Batch size (packages) & \textbackslash & 64 \\
        Training tokens (B) & 2 & 6.5 \\
        Training samples & 770k & 6m \\
        \midrule
        Auxloss weight & \multicolumn{2}{c}{0.001} \\
        SWA dropout rate & \multicolumn{2}{c}{0} \\
        \midrule
        MoBA chunk size & \multicolumn{2}{c}{512} \\
        MoBA topk & \multicolumn{2}{c}{8} \\
        \midrule
        GPU hours of A100 & 672 & 1305 \\
        \bottomrule
    \end{tabular}
    }
\end{table}

\paragraph{Knowledge Distillation}
Due to the substantial architectural modifications introduced during conversion, including MoBA and SSE modules, SpB2.0-VL starts with a relatively high training loss. To facilitate a smooth transition to the target architecture, we first perform knowledge distillation on LLaVA-NeXT-Data~\citep{liu2024llavanext}, using the original Qwen3-VL as the teacher. The objective of this stage is to rapidly recover multimodal modeling and visual understanding capability under short contexts. Following the distillation scheme used for LLM conversion, we combine layer-wise representation alignment with end-to-end logits distillation:
\begin{align}
    \mathcal{L} = \alpha \cdot 
\mathcal{L}_{\text{KD}} + \beta \cdot 
\mathcal{L}_{\text{MSE}}.
\end{align}

During VLM distillation, we set $\alpha = \beta = 1$. For the KD loss, we select the top-128 teacher logits and renormalize them to form the teacher distribution. For the MSE loss, we align the final layer outputs and average the losses across all layers.

Since VLMs require stronger global and fine-grained modeling than pure LLMs, we slightly increase the MoBA top-$k$ to 8, corresponding to 4k selected tokens, and enlarge the SWA window size to 256. We do not apply varlen packing in this stage. Training uses a global batch size of 512 on 32 A100 GPUs, with a per-device batch size of 8 and gradient accumulation of 2, at a maximum sequence length of 8k. In total, we use 770k samples (approximately 2B tokens) for distillation. After this stage, the model recovers over 95\% of the original MMStar score (49.27 vs.~51.40).

\paragraph{Instruction SFT}
After knowledge distillation, we perform instruction SFT to restore instruction-following capability. This stage is divided into two sub-stages: 2M General SFT and 4M Visual SFT. Both use data from LLaVA-OneVision-1.5-Instruct-Data~\citep{an2025llava}, but differ primarily in dataset composition. The 2M General SFT stage mixes image-text and text-only instruction data, allowing the model to jointly recover language-only and multimodal instruction-following capability. The 4M Visual SFT stage focuses on VQA-style data and further incorporates additional OCR and PixMo~\citep{deitke2025molmo} datasets to strengthen OCR and visual reasoning capability.

We use different learning rates for different model components: $5\times10^{-5}$ for the LLM backbone and projector, and $1\times10^{-5}$ for the vision encoder, together with a cosine learning-rate schedule. We also evaluate alternative settings, such as using a uniform learning rate of $1\times10^{-5}$ or $5\times10^{-5}$ for all components, and find that the adopted configuration substantially improves training stability while maintaining fast convergence. The MoBA activation ratio remains unchanged from the distillation stage.

During Instruct SFT, we apply varlen packing by grouping samples of different lengths into 16k sequences. We further use online packing, in which samples are packed dynamically during training. This avoids offline preprocessing and provides greater flexibility when adjusting dataset mixtures. In total, instruction SFT uses 6M samples, corresponding to approximately 6.5B tokens.  Training on 32 A100 GPUs requires approximately 6.8 hours per million samples.

\begin{keyfindingsbox}[Key Findings from VLM T2H Conversion]

During the development of the VLM T2H pipeline, we conducted extensive experiments and obtained several practical insights beyond routine hyperparameter tuning.

\begin{itemize}[leftmargin=1.5em,itemsep=0.35em,topsep=0.35em]
    \item \textbf{Mid-training followed by SFT is more effective than pure SFT.} Before introducing knowledge distillation, we also explored a mid-training stage based on image-caption pairs. Under the same overall data budget, adding mid-training consistently led to better capability recovery than using SFT alone. We attribute this improvement to the denser and more informative supervision provided by caption-style data.

    \item \textbf{Knowledge distillation is more effective than mid-training.} Compared with mid-training under the same data budget, knowledge distillation yields stronger capability recovery, and the advantage persists after subsequent SFT stages. We hypothesize that layer-wise distillation signals play an important role in effectively optimizing newly initialized parameters.

    \item \textbf{Learning-rate selection is critical.} In the LA layers, the normalization modules in the SSE and SWA branches are newly initialized, and their gradients are typically smaller than those of other parameters. When the learning rate is too small (e.g., $1\times10^{-5}$ or $3\times10^{-5}$), these normalization weights remain close to their initial values. Conversely, excessively large learning rates lead to unstable optimization. Empirically, we find that $5\times10^{-5}$ provides the best balance for the LLM backbone.
    
    \item \textbf{Dataset composition strongly affects capability recovery.} Increasing the proportion of OCR-related data improves OCR performance, while incorporating PixMo data enhances visual reasoning capability. More broadly, we find that introducing task-specific data in later SFT stages is an effective way to strengthen the corresponding domain capabilities.
\end{itemize}
\end{keyfindingsbox}

\section{Results}
\label{sec:results}
Based on the proposed architecture and training pipeline, we develop the SpikingBrain2.0 model family. In this section, we comprehensively evaluate the models from three perspectives: capability, efficiency, and cross-hardware compatibility.

We show that both SpB2.0-5B (Section~\ref{subsec:llm_perf}) and SpB2.0-VL-5B (Section~\ref{subsec:vlm_perf}) achieve competitive performance, effectively recovering the capability of their base Transformers at very low training cost (fewer than 7k A100 GPU hours in total).
At the same time, SpB2.0 delivers substantial efficiency gains for long-context inference, including improvements in prefilling (TTFT), decoding (TPOT), end-to-end latency, and serving throughput (Section~\ref{subsec:long_ctx_efficiency}). 
For the proposed quantization strategies, we show that both quantization paths incur only minimal performance degradation while also delivering clear efficiency benefits: the FP8 path provides practical inference speedup on H100 GPUs, whereas the INT8-Spiking path yields significant hardware-efficiency gains over the INT8 baseline (Section~\ref{subsec:quant_perf}).

\subsection{LLM Performance}
\label{subsec:llm_perf}
A primary objective of SpikingBrain2.0 is to recover the overall capability of the base Transformer at minimal additional training cost, while remaining competitive with strong open-source models of similar scale. To evaluate this goal, we assess SpB2.0-5B across three stages: the pretrained model, the instruction-tuned model, and the reasoning model.

\paragraph{Pretrained Model Performance}
To evaluate the general language modeling capability of the pretrained SpB2.0 model, we use the checkpoint obtained after the LongCT-512k stage, which has undergone only 14B tokens of continued training following conversion. We compare SpB2.0 against several strong open-source models, including Gemma3~\citep{sellergren2025medgemma}, Llama3.2~\citep{meta2024llama32}, Qwen2.5~\citep{yang2024qwen2}, and Qwen3~\citep{yang2025qwen3}. Details of the benchmark settings are provided in Appendix~\ref{app:llm_benchmarks}.

As shown in Table~\ref{tab:llm_base_comparison}, SpB2.0 achieves overall performance comparable to other Transformer baselines and remains very close to Qwen3-4B-base. Additional comparisons with the base transformer Qwen3-4B-Instruct and the previous SpB1.0-7B are provided in Appendix~\ref{app:pretrained_llm}. Notably, SpB2.0 even surpasses the base transformer model on several tasks.

\begin{table}[ht]
    \centering
    \caption{\textbf{Comparison of SpB2.0-5B with other strong open-source base models.} The best and second-best results are highlighted in bold and underline, respectively.}
    \label{tab:llm_base_comparison}
    \setlength{\tabcolsep}{6pt}
    \renewcommand{\arraystretch}{1.18}
    \resizebox{0.9\linewidth}{!}{
    \begin{tabular}{l|ccccc}
        \toprule
         & \textbf{Gemma3-4B} & \textbf{Llama3.2-3B} & \textbf{Qwen2.5-3B} & \textbf{Qwen3-4B} & \cellcolor{tabbluehead}\makecell{\textbf{SpB2.0-5B}\\\textbf{(+14B tokens)}} \\
        \midrule
        MMLU       & 59.70 & 56.71 & 65.89 & \textbf{73.09} & \cellcolor{tabblue}\underline{68.99} \\
        ARC-C      & 50.85 & 45.76 & 50.51 & \textbf{57.39} & \cellcolor{tabblue}\underline{51.53} \\
        HellaSwag  & \textbf{70.26} & \underline{69.32} & 67.71 & 67.92 & \cellcolor{tabblue}66.41 \\
        Winogrande & \textbf{72.77} & \underline{72.38} & 69.46 & 70.48 & \cellcolor{tabblue}68.35 \\
        PIQA       & \textbf{79.05} & 76.12 & \underline{78.35} & 78.18 & \cellcolor{tabblue}77.26 \\
        BBH        & 51.35 & 43.74 & 55.70 & \underline{62.11} & \cellcolor{tabblue}\textbf{65.88} \\
        GSM8K      & 47.69 & 27.85 & 75.13 & \textbf{85.22} & \cellcolor{tabblue}\underline{81.20} \\
        RULER-32k  & 43.44 & 51.38 & \underline{58.95} & \textbf{62.28} & \cellcolor{tabblue}50.87 \\
        \midrule
        \textbf{Avg.} & 59.39 & 55.41 & 65.21 & \textbf{69.58} & \cellcolor{tabblue}\underline{66.31} \\
        \bottomrule
    \end{tabular}}
\end{table}

\paragraph{Instruction-Tuned Model Performance}
To evaluate instruction-following and basic reasoning ability after SFT, we compare SpB2.0 with strong Transformer baselines as well as SpB1.0-7B-Instruct on a set of instruction-oriented generative benchmarks. We report results using the checkpoint obtained after the general SFT stage.

As shown in Table~\ref{tab:llm_gsft_comparison}, SpB2.0 achieves overall performance exceeding Qwen2.5-3B and consistently outperforms the larger SpB1.0-7B model on most benchmarks. With the exception of IFEval and MATH, its results on the remaining benchmarks stay close to those of Qwen3-4B. These results indicate that SpB2.0 retains competitive post-SFT capability despite the relatively limited training data budget used in the preceding CPT stage. We attribute the remaining gap primarily to the restricted SFT data budget and data quality, especially relative to the proprietary datasets used by other models.

\begin{table}[ht]
    \centering
    \caption{\textbf{Comparison of SpB2.0-5B with other strong open-source instruction-tuned models.} The best and second-best results are highlighted in bold and underline, respectively.}
    \label{tab:llm_gsft_comparison}
    \setlength{\tabcolsep}{6pt}
    \renewcommand{\arraystretch}{1.18}
    \resizebox{0.95\linewidth}{!}{
    \begin{tabular}{l|cccccc}
        \toprule
         & \textbf{Gemma-3-4B} & \textbf{Llama3.2-3B} & \textbf{Qwen2.5-3B} & \makecell{\textbf{Qwen3-4B}\\\textbf{(2507)}} & \textbf{SpB1.0-7B} & \cellcolor{tabbluehead}\makecell{\textbf{SpB2.0-5B}\\\textbf{(General SFT)}} \\
        \midrule
        MMLU         & 58.44 & 60.08 & 66.43 & \textbf{72.45} & 66.34 & \cellcolor{tabblue}\underline{69.06} \\
        IFEval       & \underline{75.05} & 69.50 & 58.60 & \textbf{80.96} & 43.07 & \cellcolor{tabblue}68.76 \\
        GSM8K        & \underline{84.76} & 77.71 & 74.07 & \textbf{89.16} & 66.87 & \cellcolor{tabblue}80.97 \\
        MATH         & - & 35.22 & \underline{63.16} & \textbf{72.28} & - & \cellcolor{tabblue}52.32 \\
        HumanEval    & 72.56 & 56.71 & \underline{74.39} & \textbf{86.59} & 40.24 & \cellcolor{tabblue}71.34 \\
        LongBench-32k& \textbf{41.35} & 33.73 & 28.22 & \underline{40.00} & 27.50 & \cellcolor{tabblue}37.04 \\
        \bottomrule
    \end{tabular}}
\end{table}

\paragraph{Reasoning Model Performance}
To evaluate reasoning ability with explicit chain-of-thought (CoT), we use the checkpoint after reasoning-oriented SFT (denoted as SpB2.0-5B-Thinking) and compare it against several strong Transformer baselines, including Qwen3-4B-Thinking, DeepSeek-R1-Distill-Qwen-7B, and DeepSeek-R1-Distill-Llama-8B~\citep{guo2025deepseek}. The evaluation covers a set of benchmarks known to benefit from CoT reasoning, while also including MMLU to verify that reasoning-oriented training does not cause catastrophic forgetting of general knowledge.

As shown in Table~\ref{tab:llm_rsft_comparison}, SpB2.0-5B-Thinking achieves overall performance comparable to Qwen3-Thinking and outperforms the R1-distilled models on most benchmarks, except for the more challenging AIME. We attribute the remaining gap on AIME primarily to the lack of training samples of comparable difficulty in the data mixture. 

Furthermore, we evaluate SpB2.0-5B-OPD trained via on-policy distillation. We observe that even with a limited number of training samples, the model exhibits measurable improvements in reasoning performance, highlighted by notable gains on AIME'24/25. This result provides preliminary evidence for the effectiveness of on-policy RL-style optimization on SpB2.0.

\begin{table}[ht]
    \centering
    \caption{\textbf{Comparison of SpB2.0-5B with other strong open-source reasoning models.} The best and second-best results are highlighted in bold and underline, respectively.}
    \label{tab:llm_rsft_comparison}
    \setlength{\tabcolsep}{6pt}
    \renewcommand{\arraystretch}{1.18}
    \resizebox{0.75\linewidth}{!}{
    \begin{tabular}{l|ccccc}
        \toprule
         & \makecell{\textbf{DeepSeek-}\\\textbf{R1-Distill-}\\\textbf{Qwen-7B}} & \makecell{\textbf{DeepSeek-}\\\textbf{R1-Distill-}\\\textbf{Llama-8B}} & \makecell{\textbf{Qwen3-4B-}\\\textbf{Thinking}\\\textbf{(2507)}} & \cellcolor{tabbluehead}\makecell{\textbf{SpB2.0-5B-}\\\textbf{Thinking}} & \cellcolor{tabbluehead}\makecell{\textbf{SpB2.0-5B-}\\\textbf{OPD}} \\
        \midrule
        MMLU      & 54.32 & 56.02 & \underline{68.57} & \cellcolor{tabblue}68.43 & \cellcolor{tabblue}\textbf{68.72} \\
        BBH       & 55.77 & 60.67 & \textbf{72.78} & \cellcolor{tabblue}\underline{69.16} & \cellcolor{tabblue}65.14 \\
        GSM8K     & 80.29 & 65.73 & \textbf{89.08} & \cellcolor{tabblue}83.62 & \cellcolor{tabblue}\underline{86.20} \\
        MATH      & \textbf{52.80} & 44.20 & 49.80 & \cellcolor{tabblue}\underline{52.20} & \cellcolor{tabblue}50.20 \\
        HumanEval & - & 51.22 & \underline{77.44} & \cellcolor{tabblue}75.61 & \cellcolor{tabblue}\textbf{79.27} \\
        MBPP      & 50.20 & 44.40 & \textbf{58.20} & \cellcolor{tabblue}\underline{58.00} & \cellcolor{tabblue}55.10 \\
        AIME'24   & 43.33 & 30.00 & \textbf{76.67} & \cellcolor{tabblue}33.33 & \cellcolor{tabblue}\underline{46.67} \\
        AIME'25   & 30.00 & 20.00 & \textbf{73.33} & \cellcolor{tabblue}23.33 & \cellcolor{tabblue}\underline{43.33} \\
        \bottomrule
    \end{tabular}}
\end{table}

\subsection{VLM Performance}
\label{subsec:vlm_perf}
To evaluate the multimodal capability of SpB2.0-VL-5B after instruction SFT, we conduct experiments on a comprehensive suite of multimodal benchmarks; details of the evaluation settings are provided in Appendix~\ref{app:vlm_benchmarks}. We compare SpB2.0-VL-5B with a range of widely used open-source baselines, including strong Transformer-based VLMs~\citep{Bai2025Qwen25VLTR,bai2025qwen3,chen2024far,lillava,marafiotismolvlm} as well as representative hybrid and linear architectures~\citep{li2025matvlm,zhao2025cobra,tao2025infinitevl}.

As shown in Table~\ref{tab:vlm_benchmark_comparison}, although SpB2.0-VL-5B is trained without extensive large-scale alignment, it achieves performance comparable to strong Transformer baselines such as Qwen2.5-VL-3B and LLaVA-OneVision-7B, and largely recovers the capability of the base Qwen3-VL-4B. A noticeable gap relative to the base Transformer nevertheless remains. We expect this gap to be further reduced with higher-quality domain-specific data and a larger distillation and SFT training budget.

\definecolor{tabblue}{RGB}{217, 232, 252}
\definecolor{tabbluehead}{RGB}{184, 210, 247}
\definecolor{tabgreenhead}{RGB}{216, 235, 212}
\definecolor{tabyellowhead}{RGB}{244, 239, 196}

\begin{table*}[t]
    \centering
    \caption{\textbf{Comparison of SpB2.0-VL-5B with other strong open-source VLMs on multimodal benchmarks.} Green and yellow rows denote Transformer and hybrid baselines, respectively. $^{a}$ Results are taken from the OpenVLM Leaderboard~\citep{opencompass_openvlm_leaderboard}. $^{b}$ Results are taken from the InfiniteVL~\citep{tao2025infinitevl} paper.}
    \label{tab:vlm_benchmark_comparison}
    \setlength{\tabcolsep}{5pt}
    \renewcommand{\arraystretch}{1.18}
    \resizebox{0.98\linewidth}{!}{
    \begin{tabular}{l|ccccccccc}
        \toprule
        &
        \cellcolor{tabgreenhead}\makecell{\textbf{Qwen2.5-VL-}\\\textbf{3B}} &
        \cellcolor{tabgreenhead}\makecell{\textbf{Qwen3-VL-}\\\textbf{4B}} &
        \cellcolor{tabgreenhead}\makecell{\textbf{InternVL2-}\\\textbf{4B$^{a}$}} &
        \cellcolor{tabgreenhead}\makecell{\textbf{LLaVA-}\\\textbf{OneVision-}\\\textbf{7B}} &
        \cellcolor{tabgreenhead}\makecell{\textbf{SmolVLM2-}\\\textbf{2.2B}} &
        \cellcolor{tabyellowhead}\makecell{\textbf{MaTVLM-}\\\textbf{3B$^{b}$}} &
        \cellcolor{tabyellowhead}\makecell{\textbf{Cobra-}\\\textbf{3B$^{b}$}} &
        \cellcolor{tabyellowhead}\makecell{\textbf{InfiniteVL-}\\\textbf{4B$^{b}$}} &
        \cellcolor{tabbluehead}\makecell{\textbf{SpB2.0-VL-}\\\textbf{5B}} \\
        \midrule
        AI2D                          & 80.51   & 84.00   & 79.0     & 82.51   & 69.92   & 58.9   & 46.8   & 77.2   & \cellcolor{tabblue}80.73 \\
        ChartQA                       & 84.12   & 84.44   & -      & 80.36   & 65.40   & 20.0   & 17.9   & 82.0   & \cellcolor{tabblue}77.64 \\
        DocVQA                        & 93.01   & 94.73   & -      & 86.98   & 70.75   & 33.0   & 24.0   & 91.7   & \cellcolor{tabblue}86.90 \\
        MMBench                       & 76.88   & 83.58   & 73.6   & 80.57   & 67.48   & 69.4   & 55.9   & 79.0   & \cellcolor{tabblue}78.39 \\
        MME                           & 2144.75 & 2371.74 & 2064.6 & 1987.63 & 1770.07 & 1771   & 1346   & 2126   & \cellcolor{tabblue}1981.99 \\
        MMMU(val)                     & 51.11   & 55.78   & 48.0     & 49.00   & -       & 34.4   & 31.5   & 44.0   & \cellcolor{tabblue}50.33 \\
        MMStar                        & 55.80   & 64.40   & 53.9   & 61.87   & 45.73   & 37.5   & 34.7   & 55.6   & \cellcolor{tabblue}55.40 \\
        OCRBench                      & 82.60   & 84.50   & 78.4   & 63.50   & 67.70   & 35.1   & 30.7   & 79.8   & \cellcolor{tabblue}75.10 \\
        RealWorldQA                   & 65.10   & 71.90   & 60.5   & 69.54   & 56.99   & 52.3   & 51.0   & 67.3   & \cellcolor{tabblue}67.19 \\
        SEEDBench$_{\text{IMG}}$                & 73.88   & 78.46   & 73.2   & 76.64   & 70.69   & 65.6   & 63.3   & 72.9   & \cellcolor{tabblue}76.08 \\
        ScienceQA                     & 79.54   & 93.42   & 96.3   & 94.71   & 86.60   & 65.9   & 69.3   & 93.4   & \cellcolor{tabblue}81.55 \\
        TextVQA                       & 79.29   & 81.86   & -      & 75.92   & 68.81   & 53.2   & 42.9   & 78.5   & \cellcolor{tabblue}71.72 \\
        \bottomrule
    \end{tabular}}
\end{table*}

\subsection{Long-context Efficiency}
\label{subsec:long_ctx_efficiency}
In this section, we evaluate the long-context serving efficiency of SpB2.0. The results show that SpB2.0 delivers substantial efficiency gains while maintaining competitive capability under a very low training budget. We consider two widely used deployment settings: HuggingFace sequence parallelism (Section~\ref{subsubsec:efficiency_hf_sp}) and vLLM tensor parallelism (Section~\ref{subsubsec:efficiency_vllm_tp}). We report multiple serving metrics, including prefilling latency (time-to-first-token, TTFT), decoding latency (time-per-output-token, TPOT), end-to-end latency, and throughput.

\subsubsection{Speedup under HuggingFace Sequence Parallelism}
\label{subsubsec:efficiency_hf_sp}
We first evaluate SpB2.0-5B under a standard HuggingFace setup with full sequence parallelism (SP), and compare its TTFT against the baseline Qwen3-4B, since TTFT is a key metric for real-world user experience.

We fix the batch size to 1 and proportionally scale both the input sequence length and the SP degree, increasing the sequence length from 250k to 4M tokens while maintaining a constant per-GPU token budget of 125k. Both models use Ulysses for all-to-all communication. All experiments are conducted on H100 80GB GPUs.

As shown in Table~\ref{tab:hf_sp_speedup}, SpB2.0 achieves a 10.13$\times$ TTFT speedup over the base Transformer at 4M context length. The observed scaling trend is consistent with theoretical complexity. For full-attention-based Qwen3, quadratic attention cost leads to approximately linear growth in TTFT as both sequence length and GPU count increase proportionally. In contrast, the hybrid attention in SpB2.0 exhibits sub-quadratic compute growth and correspondingly sub-linear TTFT scaling. As a result, the relative TTFT advantage of SpB2.0 becomes larger at longer context lengths, making it particularly suitable for ultra-long-context serving.

\begin{table}[t]
    \centering

    \begin{minipage}[t]{0.4\textwidth}
        \centering
        \captionof{table}{\textbf{TTFT comparison between Qwen3-4B and SpB2.0-5B under HuggingFace sequence parallelism.}}
        \label{tab:hf_sp_speedup}
        \setlength{\tabcolsep}{7pt}
        \renewcommand{\arraystretch}{1.18}
        \resizebox{\linewidth}{!}{
        \begin{tabular}{cc|cc|c}
            \toprule
            \textbf{Length} & \makecell{\textbf{GPUs}\\\textbf{(SP Degree)}} & \textbf{Qwen3} & \textbf{SpB2.0} & \textbf{Speedup} \\
            \midrule
            250K & 2  & 28932  & 15734 & 1.84$\times$ \\
            500K & 4  & 57314  & 18290 & 3.13$\times$ \\
            1M   & 8  & 115121 & 22950 & 5.02$\times$ \\
            2M   & 16 & 236265 & 33100 & 7.14$\times$ \\
            4M   & 32 & 470505 & 46455 & 10.13$\times$ \\
            \bottomrule
        \end{tabular}}
    \end{minipage}
    \hfill
    \begin{minipage}[t]{0.55\textwidth}
        \centering
        \captionof{table}{\textbf{Serving throughput comparison between SpB2.0-5B and Qwen3-4B under vLLM tensor parallelism.}}
        \label{tab:vllm_throughput}
        \setlength{\tabcolsep}{7pt}
        \renewcommand{\arraystretch}{1.18}
        \resizebox{\linewidth}{!}{
        \begin{tabular}{l|ccc}
            \toprule
             & \makecell{\textbf{Prompt Throughput}\\\textbf{(tokens/s)}} & \makecell{\textbf{Generation Throughput}\\\textbf{(tokens/s)}} & \makecell{\textbf{Concurrent}\\\textbf{Requests}} \\
            \midrule
            \textbf{SpB2.0-5B} & 48671.09 & 47.53 & 76 \\
            \textbf{Qwen3-4B}  & 30927.56 & 30.20 & 24 \\
            \midrule
            \textbf{Speedup}   & 1.57$\times$ & 1.57$\times$ & 3.17$\times$ \\
            \bottomrule
        \end{tabular}}
    \end{minipage}
\end{table}

\subsubsection{Speedup under vLLM Tensor Parallelism}
\label{subsubsec:efficiency_vllm_tp}
\paragraph{TTFT, TPOT, and End-to-End Latency} Given the extensive system-level optimizations in vLLM for practical deployment, we further evaluate SpB2.0 under this framework and compare its serving efficiency with Qwen3-4B. All experiments are conducted on 8 A100 80GB GPUs with tensor parallelism (TP = 8).

As summarized in Table~\ref{tab:vllm_latency}, the efficiency advantage of SpB2.0 continues to increase with sequence length. Without chunked prefill, when the prompt length increases from 128k to 512k tokens with the output length fixed at 100 tokens, SpB2.0 achieves progressively larger gains. At 512k context length, it delivers 4.5$\times$, 1.12$\times$, and 4.3$\times$ speedups in TTFT, TPOT, and end-to-end latency, respectively. For prompt lengths beyond 512k, we enable chunked prefill due to memory constraints. Even at 2M context length, SpB2.0 still provides substantial acceleration across all three metrics. Notably, beyond 4M tokens, Qwen3 exceeds memory limits, whereas SpB2.0 continues to support serving at lengths above 10M tokens.

\begin{table}[t]
    \centering
    \caption{\textbf{Long-context inference latency comparison between SpB2.0-5B and Qwen3-4B under vLLM tensor parallelism.} TTFT denotes time to first token, TPOT denotes time per output token, and End2End denotes total generation latency. Qwen3-4B exceeds GPU memory limits beyond 4M tokens.}
    \label{tab:vllm_latency}
    \setlength{\tabcolsep}{6pt}
    \renewcommand{\arraystretch}{1.18}
    \resizebox{0.8\linewidth}{!}{
    \begin{tabular}{l|rrr|rrr}
        \toprule
        & \multicolumn{3}{c|}{\textbf{SpB2.0-5B}} & \multicolumn{3}{c}{\textbf{Qwen3-4B}} \\
        \midrule
        \textbf{Prompt Length} & \textbf{TTFT (s)} & \textbf{TPOT (s)} & \textbf{End2End (s)} & \textbf{TTFT (s)} & \textbf{TPOT (s)} & \textbf{End2End (s)} \\
        \midrule
        128K & 3.0947 & 0.0090 & 3.9857 & 4.9580 & 0.0063 & 5.5835 \\
        256K & 6.5905 & 0.0093 & 7.5113 & 17.8807 & 0.0080 & 18.6771 \\
        512K & 15.0288 & 0.0105 & 15.9903 & 67.5043 & 0.0118 & 68.6767 \\
        \midrule
        \multicolumn{7}{l}{\textbf{Enable chunked prefill for longer context}} \\
        \midrule
        1M  & 120.5183 & 0.0119 & 121.6955 & 247.5864 & 0.0146 & 249.0308 \\
        2M  & 368.2790 & 0.0148 & 369.7411 & 953.9603 & 0.0251 & 956.4483 \\
        4M  & 1250.8069 & 0.0206 & 1252.8452 & \multicolumn{3}{c}{OOM} \\
        8M  & 4582.5763 & 0.0332 & 4585.8591 & \multicolumn{3}{c}{OOM} \\
        10M & 6567.2947 & 0.0467 & 6571.9159 & \multicolumn{3}{c}{OOM} \\
        \bottomrule
    \end{tabular}}
\end{table}

Compared with Qwen3, SpB2.0 also introduces several sources of overhead, including additional QKV projections in LA-SWA layers and extra TP communication from merge norm. As a result, decoding speedups become more pronounced only at longer context lengths. In addition, under chunked prefill, the measured speedup remains below the theoretical bound due to current limitations in MoBA paged-attention implementations. We expect further engineering optimization to unlock additional efficiency gains.

\paragraph{Throughput Improvement}
Leveraging the reduced memory footprint of the hybrid architecture, we further evaluate serving throughput under the vLLM framework. The experiment uses 128k input tokens and 128 output tokens. As shown in Table~\ref{tab:vllm_throughput}, SpB2.0 supports 3.17$\times$ higher request concurrency due to its lower KV-cache memory usage, resulting in a 1.57$\times$ improvement in prompt and generation throughput. With further optimization of the vLLM implementation, the concurrency gain is expected to move closer to the theoretical upper bound of 4$\times$.

\subsection{Quantization Performance}
\label{subsec:quant_perf}
\subsubsection{Capability Preservation under Quantization}
\label{subsubsec:capability_perf_quant}
We evaluate the pretrained SpikingBrain2.0-5B model on eight general benchmarks to assess the impact of quantization on model capability. As shown in Table~\ref{tab:quantization_perf_benchmark}, the Qwen3-4B baseline achieves an average score of 0.6588, while full-precision SpB2.0 attains a competitive average score of 0.6392. Among the quantized variants, the INT8 model (SpB2.0-AINT8-WINT8) achieves an average score of 0.6348, corresponding to only a 0.69\% performance drop relative to full-precision SpB2.0. Measurements show that the firing rate of its spike sequences is merely 35.69\% (i.e., a spike sequence sparsity of 64.31\%). The FP8 model (SpB2.0-AFP8-WFP8) further reduces the performance drop, reaching an average score of 0.6377 with only a 0.24\% drop. These results show that both quantization paths preserve the core capability of SpikingBrain2.0 with minimal degradation. In particular, the INT8 path validates the feasibility of neuromorphic-oriented deployment, while the FP8 path provides a practical solution for efficient deployment on modern GPUs.

\begin{table}[t]
    \centering
    \caption{\textbf{Quantization comparison of SpB2.0 on language benchmarks.} We compare the full-precision model with its INT8 and FP8 variants across eight language benchmarks. Avg. denotes the average score over all benchmarks.}
    \label{tab:quantization_perf_benchmark}
    \setlength{\tabcolsep}{6pt}
    \renewcommand{\arraystretch}{1.15}
    \resizebox{0.98\linewidth}{!}{
    \begin{tabular}{l|cccccccc|c}
        \toprule
        & \textbf{ARC-E} & \textbf{ARC-C} & \textbf{HellaSwag} & \textbf{BoolQA} & \textbf{OpenbookQA} & \textbf{PiQA} & \textbf{Winogrande} & \textbf{MMLU} & \textbf{Avg.} \\
        \midrule
        \textbf{Qwen3-4B}                    & 0.8043 & 0.5384 & 0.6854 & 0.8502 & 0.3000   & 0.7492 & 0.6598 & 0.6834 & 0.6588 \\
        \textbf{SpB2.0}            & 0.8144 & 0.5273 & 0.6771 & 0.7627 & 0.3040 & 0.7617 & 0.6417 & 0.6244 & 0.6392 \\
        \textbf{SpB2.0-AINT8-WINT8}& 0.8116 & 0.5188 & 0.6758 & 0.7489 & 0.3040 & 0.7617 & 0.6354 & 0.6218 & 0.6348 \\
        \textbf{SpB2.0-AFP8-WFP8}  & 0.8144 & 0.5245 & 0.6740  & 0.7572 & 0.3100  & 0.7639 & 0.6353 & 0.6219 & 0.6377 \\
        \bottomrule
    \end{tabular}}
\end{table}

\subsubsection{Efficiency Gains from Quantization}
\label{subsubsec:efficiency_perf_quant}
\paragraph{FP8 Format}
The FP8 quantization path provides practical inference acceleration on NVIDIA Hopper GPUs, making it a viable deployment option on existing hardware. To evaluate this benefit, we compare SpB2.0-AFP8-WFP8 with its full-precision counterpart under the HuggingFace sequence-parallel (SP) setting, using TTFT as the primary metric across different input lengths. As shown in Table~\ref{tab:quant_sp_speedup}, the FP8 variant consistently achieves substantial speedup over the BF16 baseline at all tested sequence lengths. In particular, under the 250k (SP=2) setting, the TTFT speedup reaches 2.52$\times$. After FP8 quantization, SpB2.0 achieves up to 15.13$\times$ TTFT speedup over full-precision Qwen3-4B at 4M context length\footnote{Qwen3 can also benefit from FP8 quantization. Therefore, this comparison reflects the combined advantage of the SpB2.0 architecture and the FP8 inference path, rather than the effect of quantization alone.}.

\begin{table}[t]
    \centering
    \caption{\textbf{TTFT comparison between full-precision SpB2.0-BF16 and SpB2.0-AFP8-WFP8.}}
    \label{tab:quant_sp_speedup}
    \setlength{\tabcolsep}{8pt}
    \renewcommand{\arraystretch}{1.18}
    \resizebox{0.55\linewidth}{!}{
    \begin{tabular}{cc|cc|c}
        \toprule
        \textbf{Length} & \makecell{\textbf{GPUs}\\\textbf{(SP Degree)}} & \makecell{\textbf{SpB2.0-}\\\textbf{BF16}} & \makecell{\textbf{SpB2.0-}\\\textbf{AFP8-WFP8}} & \textbf{Speedup} \\
        \midrule
        250K & 2  & 15734 & 6248  & 2.52$\times$ \\
        500K & 4  & 18290 & 17772 & 1.03$\times$ \\
        1M   & 8  & 22950 & 20442 & 1.12$\times$ \\
        2M   & 16 & 33100 & 25463 & 1.30$\times$ \\
        4M   & 32 & 46455 & 31095 & 1.49$\times$ \\
        \bottomrule
    \end{tabular}}
\end{table}

\paragraph{INT8–Spiking Format}
We implement a hardware prototype of the proposed architecture in Section~\ref{subsubsec:int_spike_quant}, which reads weights and input data from external interfaces, performs tiled Integer–Spiking matrix multiplication, and outputs final results. For comparison, we also implement an INT8 matrix-multiplication baseline. Both designs are synthesized with 28nm technology using Synopsys Design Compiler to obtain area and timing metrics. We then perform gate-level simulation using post-synthesis netlists and real network inputs, and annotate switching activities to estimate power under realistic workloads. For a fair comparison, area is reported using the 500MHz synthesis netlist, while power is reported at both 250MHz and 500MHz.

\begin{table}[ht]
    \centering
    \caption{\textbf{Hardware area and power comparison between the INT8 baseline and the proposed Integer-Spiking design.}}
    \label{tab:spiking_hardware_comparison}
    \setlength{\tabcolsep}{8pt}
    \renewcommand{\arraystretch}{1.18}
    \resizebox{0.7\linewidth}{!}{
    \begin{tabular}{l|cc|c}
        \toprule
        \textbf{Metric} & \textbf{INT8 Baseline} & \textbf{Integer-Spiking} & \textbf{Normalized Ratio} \\
        \midrule
        Power @250MHz (mW) & 310 & 161 & 0.519$\times$ \\
        Power @500MHz (mW) & 593 & 317 & 0.535$\times$ \\
        Area ($\mu\text{m}^2$) & 2,168,626 & 636,541 & 0.294$\times$ \\
        \bottomrule
    \end{tabular}}
\end{table}

As shown in Table~\ref{tab:spiking_hardware_comparison}, the Integer–Spiking implementation based on the Adaptive Spiking Coding scheme and dedicated hardware architecture provides stable and significant hardware-efficiency gains over the INT8 baseline: area is reduced by 70.6\%, and power is reduced by 48.1\% and 46.5\% at 250MHz and 500MHz, respectively (corresponding normalized ratios: 0.294$\times$, 0.519$\times$, and 0.535$\times$). These results indicate strong potential for resource-constrained edge deployments.


\section{Conclusion}
In this paper, we present SpikingBrain2.0, a hybrid spiking foundation model family designed to improve long-context efficiency with minimal additional training cost. SpB2.0 combines an inter-layer hybrid attention architecture termed Dual-Space Sparse Attention (DSSA), integrating MoBA and SSE, with hardware-oriented activation coding to address the dominant computational bottlenecks across different sequence-length regimes and deployment platforms. To support efficient architectural migration, we further develop optimized Transformer-to-Hybrid conversion pipelines for both LLMs and VLMs using only open-source data.

Extensive experiments show that SpB2.0-5B and SpB2.0-VL-5B recover most of the capability of their base Transformers while delivering substantial long-context inference gains, including strong improvements in TTFT, TPOT, end-to-end latency, serving throughput, and maximum supported context length. In addition, the proposed quantization strategies enable efficient deployment across heterogeneous hardware platforms, including modern GPUs and neuromorphic hardware, with only minimal performance degradation.

Overall, SpikingBrain2.0 demonstrates that sparse-memory design, efficient hybrid attention, spiking coding, and lightweight conversion training can be effectively combined in a unified foundation-model framework. We hope this work provides a practical step toward efficient multimodal large models for long-context, edge, and resource-constrained applications.

\clearpage
\bibliography{main}
\bibliographystyle{tmlr}

\clearpage
\appendix

\section{Notations}\label{app:notations}

\begin{table}[ht]
\centering
\caption{\textbf{Notations used in this work.}}
\small
\vspace{-1.5\baselineskip}
\setlength{\extrarowheight}{1pt}
\begin{longtable}{c|L{10cm}}
\toprule
Symbol & Description \\
\midrule
$\mathbf{x}_t$ &
	Row-vector representation at time step $t$. \\
$\mathbf{X}_t$ &
	Matrix representation at time step $t$, or the $t$-th block in MoBA. \\
$\mathbf{M}$ &
	Causal mask, where $\mathbf{M}_{ij}=1$ if $i\geq j$ and $\mathbf{M}_{ij}=0$ otherwise. \\
$\mathbf{e}_t^i$ &
	The $i$-th element of vector $\mathbf{e}_t$. \\
$\mathbf{S}_t^i$ &
	The $i$-th partition of state matrix $\mathbf{S}_t$. \\
\midrule
$n$ &
	Sequence length. \\
$t,s$ &
	Time-step indices. \\
$N$ &
	Number of sparse state partitions in SSE. \\
$k$ &
	Top-$k$ value used in MoBA and SSE. \\
$b$ &
	Block size in MoBA. \\ 
$w$ &
	Window size in SWA. \\
\bottomrule
\end{longtable}
\end{table}

\section{Definitions of Loss Terms Used in T2H Conversion}\label{app:loss_def}
In addition to the standard cross-entropy (CE) objective, several auxiliary loss terms are used during T2H conversion.

\paragraph{Auxiliary Loss for SSE Layers.}
To improve training stability and mitigate partition-selection imbalance during sparse state updates, we apply an auxiliary loss to SSE layers:
\begin{align}
    \mathcal{L}_{\text{aux}}=\frac{N}{k} \sum_{t=1}^n\sum_{i=1}^N \mathbf{f}_t^i \mathbf{e}_t^i,
\end{align}
where $\mathbf{e}_t^i$ denotes the gating value of the $i$-th partition at token $t$, and $\mathbf{f}_t^i$ denotes its corresponding selection frequency. The auxiliary losses from all linear-attention layers are summed to obtain the final $\mathcal{L}_{\text{aux}}$.

\paragraph{Layer-wise MSE Alignment Loss.}
To encourage representation alignment between the student and teacher models, we introduce a layer-wise mean squared error (MSE) loss. For each layer $l$, the student representation $\mathbf{h}^{\text{student}}_{t,l}$ is aligned with the corresponding teacher representation $\mathbf{h}^{\text{teacher}}_{t,l}$:
\begin{align}
    \mathcal{L}_{\text{MSE},l} = \frac{1}{n} \sum_{t=1}^{n}  ||\mathbf{h}^{\text{student}}_{t,l} - \mathbf{h}^{\text{teacher}}_{t,l} ||^2. 
\end{align}
The losses across all layers are then averaged to produce the final alignment loss $\mathcal{L}_{\text{MSE}}$. For LLM distillation, we align the attention outputs of each layer. For VLM distillation, we instead align the final outputs at each layer.

\paragraph{End-to-End Knowledge Distillation Loss.}
We further apply logits-level knowledge distillation by minimizing the KL divergence between the teacher and student predictions. For each token, only the top-$K$ teacher logits are considered:
\begin{align}
    \mathcal{L}_{\text{KD}} = \frac{1}{n}\sum_{i=1}^{n}\left(\sum_{k=1}^{K} p_k^{\text{teacher}} \log \frac{p_k^{\text{teacher}}}{p_k^{\text{student}}}\right).
\end{align}
This formulation concentrates the distillation signal on the most informative logits while reducing computational overhead.

\section{Experiments}
\label{app:experiments}

\subsection{LLM Benchmarks}
\label{app:llm_benchmarks}
Tables~\ref{tab:bmk_setting_llm_base} and~\ref{tab:bmk_setting_llm_ins} present the benchmark settings and evaluation frameworks used for pretrained and instruction-tuned LLMs, respectively.

\begin{table}[htbp]
    \centering
    \caption{\textbf{Benchmark settings and evaluation frameworks for pretrained language models.}}
    \label{tab:bmk_setting_llm_base}
    \begin{tabular}{l|l|l}
        \toprule
        \textbf{Benchmark} & \textbf{Settings} & \textbf{Framework} \\
        \midrule
        \textbf{General} & & \\
        \quad MMLU ~\citep{hendryckstest2021mmlu} & logprobs, 5-shot & \texttt{opencompass} \\
        \quad ARC-C ~\citep{allenai:arc} & logprobs, 5-shot & \texttt{opencompass} \\
        \quad HellaSwag ~\citep{zellers2019hellaswag} & logprobs, 0-shot & \texttt{opencompass} \\
        \quad Winogrande ~\citep{sakaguchi2019winogrande} & logprobs, 5-shot & \texttt{opencompass} \\
        \quad PIQA ~\citep{Bisk2020PIQA} & logprobs, 0-shot & \texttt{lm-eval-harness} \\
        \quad BBH  ~\citep{suzgun2022challengingbbh} & strict match, 3-shot CoT & \texttt{opencompass} \\
        
        \midrule
        \textbf{Math} & & \\
        \quad GSM8k ~\citep{cobbe2021gsm8k} & strict match, 4-shot CoT & \texttt{opencompass} \\
        \midrule
        \textbf{Long Context} & & \\
        \quad RULER-32k ~\citep{hsieh2024ruler} & strict match & \texttt{opencompass} \\
        \bottomrule
    \end{tabular}
\end{table}

\begin{table}[htbp]
    \centering
    \caption{\textbf{Benchmark settings and evaluation frameworks for instruction-tuned language models.}}
    \label{tab:bmk_setting_llm_ins}
    \begin{tabular}{l|l|l}
        \toprule
        \textbf{Benchmark} & \textbf{Settings} & \textbf{Framework} \\
        \midrule
        \textbf{General} & & \\
        \quad MMLU & logprobs, 5-shot & \texttt{opencompass} \\

        \midrule
        \textbf{Instruction Following} & & \\
        \quad IFEVal ~\citep{zhou2023ifeval} & strict prompt accuracy & \texttt{opencompass} \\
        \midrule
        \textbf{Math} & & \\
        \quad GSM8k & strict match, 4-shot CoT & \texttt{opencompass} \\
        \quad MATH  ~\citep{hendrycksmath2021math} & strict match, 4-shot CoT & \texttt{opencompass} \\
        \midrule
        \textbf{Code} & & \\
        \quad HumanEval ~\citep{chen2021codexhumaneval} & pass@1 & \texttt{opencompass} \\
        \midrule
        \textbf{Long Context} & & \\
        \quad LongBench-32k ~\citep{bai2024longbench} & strict match & \texttt{opencompass} \\
        \bottomrule
    \end{tabular}
\end{table}

\subsection{VLM Benchmarks}
\label{app:vlm_benchmarks}
Table~\ref{tab:bmk_setting_vlm} summarizes the benchmark settings and evaluation frameworks used for VLMs. For all benchmarks, to ensure reliable evaluation, we first apply rule-based matching to extract the predicted answer; if no valid answer is identified, we then use a judge model for verification. The judge model is a locally deployed Qwen3-VL-2B. To mitigate the tendency of Qwen3-VL to produce overly long outputs that complicate evaluation, we use the following system prompt during evaluation: \texttt{system\_prompt="When answering the questions, do NOT output any explanations, thinking process or intermediate steps, you should ONLY give the answer."} For fairness, the same system prompt is also applied to SpB2.0-VL.

\begin{table}[htbp]
    \centering
    \caption{\textbf{Benchmark settings and evaluation frameworks for VLMs.}}
    \label{tab:bmk_setting_vlm}
    \begin{tabular}{l|l|l}
        \toprule
        \textbf{Benchmark} & \textbf{Settings} & \textbf{Framework} \\
        \midrule
        AI2D ~\citep{kembhavi2016diagram}& MCQ & \texttt{VLMEvalKit} \\
        ChartQA ~\citep{masry2022chartqa}& VQA & \texttt{VLMEvalKit} \\
        DocVQA ~\citep{mathew2021docvqa}& VQA & \texttt{VLMEvalKit} \\
        MMBench ~\citep{liu2024mmbench}& MCQ & \texttt{VLMEvalKit} \\
        MME ~\citep{fumme}& Y/N & \texttt{VLMEvalKit} \\
        MMMU(val) ~\citep{yue2024mmmu}& MCQ & \texttt{VLMEvalKit} \\
        MMStar ~\citep{chen2024we}& MCQ & \texttt{VLMEvalKit} \\
        OCRBench ~\citep{liu2024ocrbench}& VQA & \texttt{VLMEvalKit} \\
        RealWorldQA \citep{xai2024realworldqa}& MCQ & \texttt{VLMEvalKit} \\
        SEEDBench$_{\text{IMG}}$ ~\citep{li2024seed}& MCQ & \texttt{VLMEvalKit} \\
        ScienceQA ~\citep{saikh2022scienceqa}& MCQ & \texttt{VLMEvalKit} \\
        TextVQA ~\citep{singh2019towards}& VQA & \texttt{VLMEvalKit} \\
        \bottomrule
    \end{tabular}
\end{table}

\subsection{Additional Pretrained Model Comparisons}
\label{app:pretrained_llm}
As shown in shown in Table~\ref{tab:llm_base_comparison_full}, in addition to the strong transformer baselines, we further compare the pretrained SpB2.0-5B model with the base transformer model used for T2H conversion, Qwen3-4B-Instruct, as well as the previous SpB1.0-7B. Notably, despite using only 14B open-source tokens for continual training, SpB2.0 surpasses the base Transformer on several tasks, including HellaSwag, Winogrande, and PIQA. It also outperforms the larger SpB1.0-7B on several challenging benchmarks, such as MMLU, BBH, GSM8K, and RULER.

\begin{table}[ht]
    \centering
    \caption{\textbf{Additional comparison of the pretrained SpB2.0-5B model with Qwen3-4B-Instruct and SpB1.0-7B.} Qwen3-4B-Instruct, as an instruction-tuned model, performs well on generative benchmarks such as BBH, GSM8K, and RULER-32k. SpB1.0-7B, as a pure linear-complexity model with a fixed state size, performs poorly on RULER-32k.}
    \label{tab:llm_base_comparison_full}
    \setlength{\tabcolsep}{6pt}
    \renewcommand{\arraystretch}{1.18}
    \resizebox{0.98\linewidth}{!}{
    \begin{tabular}{l|ccccccc}
        \toprule
         & \textbf{Gemma3-4B} & \textbf{Llama3.2-3B} & \textbf{Qwen2.5-3B} & \makecell{\textbf{Qwen3-4B-}\\\textbf{Base}} & \textbf{SpB1.0-7B} & \makecell{\textbf{Qwen3-4B-}\\\textbf{Instruct-2507}} & \cellcolor{tabbluehead}\textbf{SpB2.0-5B} \\
        \midrule
        MMLU      & 59.70 & 56.71 & 65.89 & 73.09 & 65.80 & 72.45 & \cellcolor{tabblue}68.99 \\
        ARC-C     & 50.85 & 45.76 & 50.51 & 57.39 & 52.54 & 58.98 & \cellcolor{tabblue}51.53 \\
        HellaSwag & 70.26 & 69.32 & 67.71 & 67.92 & 72.97 & 64.20 & \cellcolor{tabblue}66.41 \\
        Winogrande& 72.77 & 72.38 & 69.46 & 70.48 & 73.48 & 67.80 & \cellcolor{tabblue}68.35 \\
        PIQA      & 79.05 & 76.12 & 78.35 & 78.18 & 80.03 & 76.01 & \cellcolor{tabblue}77.26 \\
        BBH       & 51.35 & 43.74 & 55.70 & 62.11 & 53.30 & 80.05 & \cellcolor{tabblue}65.88 \\
        GSM8K     & 47.69 & 27.85 & 75.13 & 85.22 & 68.08 & 87.95 & \cellcolor{tabblue}81.20 \\
        RULER-32k & 43.44 & 51.38 & 58.95 & 62.28 & 17.21 & 85.62 & \cellcolor{tabblue}50.87 \\
        \midrule
        \textbf{Avg.} & 59.39 & 55.41 & 65.21 & 69.58 & 60.43 & 74.13 & \cellcolor{tabblue}66.31 \\
        \bottomrule
    \end{tabular}}
\end{table}

\subsection{Extended Results for Long-Context Extension}
\label{app:longct_perf}
To better understand the effect of progressive long-context training, we evaluate intermediate checkpoints of SpB2.0-5B throughout the long-context continual-training pipeline. As shown in Table~\ref{tab:longct_stage_comparison}, performance on long-context benchmarks, measured by RULER-80k, improves steadily across the three LongCT stages, while the short-context benchmark MMLU decreases only slightly.

\begin{table}[ht]
    \centering
    \caption{\textbf{Performance across different stages of long-context extension on RULER-80k and MMLU.} Long-context performance improves steadily from LongCT-stage1 to LongCT-stage3, while MMLU decreases only slightly. SN: single-needle; MK: multi-key; MQ: multi-query; MV: multi-value; SQD: SQuAD.}
    \label{tab:longct_stage_comparison}
    \setlength{\tabcolsep}{8pt}
    \renewcommand{\arraystretch}{1.2}
    \resizebox{0.95\linewidth}{!}{
    \begin{tabular}{l|ccccccc|c}
        \toprule
         & \textbf{SN1} & \textbf{SN2} & \textbf{SN3} & \textbf{MK1} & \textbf{MQ} & \textbf{MV} & \textbf{SQD} & \textbf{MMLU} \\
        \midrule
        \textbf{LongCT-stage1-64k}  & 96.20 & 58.00 & 23.80 & 34.60 & 23.50 & 44.20 & 17.78 & \textbf{69.36} \\
        \textbf{LongCT-stage2-256k} & 99.80 & 93.40 & 83.60 & 61.40 & 71.60 & 78.20 & 25.53 & 69.04 \\
        \textbf{LongCT-stage3-512k} & \textbf{100.0} & \textbf{94.40} & \textbf{89.00} & \textbf{68.60} & \textbf{77.85} & \textbf{85.45} & \textbf{26.47} & 68.99 \\
        \bottomrule
    \end{tabular}}
\end{table}

\subsection{Ablation on the SSE-SWA Design}
\label{app:sse_swa_ablation}
In the design of the SSE-SWA block, our goal is to reduce interference from the auxiliary SWA branch with the main LA pathway and to avoid computational shortcuts that may weaken LA learning. To verify this design, we compare the trained SpB2.0 checkpoint with two ablated inference variants, in which only the SSE branch or only the SWA branch is enabled. As shown in Table~\ref{tab:sse_swa_ablation}, the combination of a small SWA window, branch dropout, and merge norm effectively preserves the LA branch. In particular, SSE-only inference retains most of the capability of the full model, whereas SWA-only inference leads to a substantial performance drop.

\begin{table}[ht]
    \centering
    \caption{\textbf{Ablation of SSE-only and SWA-only inference variants.} All models are evaluated from the SpB2.0 checkpoint obtained after short-context distillation on 400k samples. For ablation, inference is performed with only the SSE branch or only the SWA branch enabled.}
    \label{tab:sse_swa_ablation}
    \setlength{\tabcolsep}{8pt}
    \renewcommand{\arraystretch}{1.18}
    \resizebox{0.6\linewidth}{!}{
    \begin{tabular}{l|ccc}
        \toprule
        & \textbf{SpB2.0} & \makecell{\textbf{SpB2.0-}\\\textbf{SSE-only}} & \makecell{\textbf{SpB2.0-}\\\textbf{SWA-only}} \\
        \midrule
        MMLU      & 69.38 & 62.15 & 23.69 \\
        ARC-C    & 52.88 & 52.54 & 25.76 \\
        HellaSwag & 66.26 & 62.78 & 31.41 \\
        \bottomrule
    \end{tabular}}
\end{table}

\end{document}